%% file: main.tex
\pdfoutput=1

\documentclass[11pt]{article}

\usepackage[final]{acl}

\usepackage{times}
\usepackage{latexsym}

\usepackage[T1]{fontenc}

\usepackage[utf8]{inputenc}

\usepackage{microtype}

\usepackage{inconsolata}

\usepackage{multirow}           
\usepackage{booktabs}           
\usepackage{arydshln}
\usepackage{graphicx}           
\usepackage{float}              
\usepackage{bm}                 
\usepackage[tikz]{bclogo}       
\usepackage{amsmath}            
\usepackage{amsfonts}           
\usepackage{pgfplots}           
\pgfplotsset{compat=1.18}
\usepackage{caption}
\usepackage{subcaption}         
\usepackage{enumitem}           

\title{CLEME2.0: Towards Interpretable Evaluation by Disentangling Edits\\for Grammatical Error Correction}

\author{
Jingheng Ye$^{1}$\thanks{Equal Contribution.},
Zishan Xu$^{1}$\footnotemark[1],
Yinghui Li$^{1}$,
Linlin Song$^{2}$,
Qingyu Zhou$^{3}$,
Hai-Tao Zheng$^{1,4}$\thanks{Corresponding Author: Hai-Tao Zheng. (E-mail: zheng.haitao@sz.tsinghua.edu.cn)},\\
\textbf{Ying Shen}$^{5}$,
\textbf{Wenhao Jiang}${^6}$,
\textbf{Hong-Gee Kim}${^7}$,
\textbf{Ruitong Liu}${^1}$,
\textbf{Xin Su}${^8}$,
\textbf{Zifei Shan}${^8}$\\
$^{1}$Tsinghua University,
$^{2}$Huazhong University of Science and Technology,\\
$^{3}$ByteDance Inc.,
$^{4}$Peng Cheng Laboratory,
$^{5}$Sun-Yat Sen University,\\
$^{6}$Guangdong Laboratory of Artificial Intelligence and Digital Economy (SZ),\\
$^{7}$Seoul National University,
$^{8}$Tencent\\
\texttt{\{yejh22,xzs23\}@mails.tsinghua.edu.cn}
}

\begin{document}
\maketitle

\definecolor{brickred}{HTML}{b92622}
\definecolor{midnightblue}{HTML}{005c7f}
\definecolor{salmon}{HTML}{f1958d}
\definecolor{burntorange}{HTML}{f19249}
\definecolor{junglegreen}{HTML}{4dae9d}
\definecolor{forestgreen}{HTML}{499c5e}
\definecolor{pinegreen}{HTML}{3d8a75}
\definecolor{seagreen}{HTML}{6bc1a2}
\definecolor{limegreen}{HTML}{97c65a}
\newcommand{\white}[1]{\textcolor{white}{#1}}
\newcommand{\brickred}[1]{\textcolor{brickred}{#1}}
\newcommand{\midnightblue}[1]{\textcolor{midnightblue}{#1}}
\newcommand{\salmon}[1]{\textcolor{salmon}{#1}}
\newcommand{\junglegreen}[1]{\textcolor{junglegreen}{#1}}
\newcommand{\forestgreen}[1]{\textcolor{forestgreen}{#1}}
\newcommand{\pinegreen}[1]{\textcolor{pinegreen}{#1}}
\newcommand{\seagreen}[1]{\textcolor{seagreen}{#1}}

\newcommand{\MetricName}{CLEME2.0}
\newcommand{\TP}[1]{\textcolor{forestgreen}{#1}}
\newcommand{\FP}[1]{\textcolor{magenta}{#1}}
\newcommand{\FPne}[1]{\textcolor{blue}{#1}}
\newcommand{\FPun}[1]{\textcolor{orange}{#1}}
\newcommand{\FN}[1]{\textcolor{olive}{#1}}
\newcommand{\TODO}[1]{\textcolor{blue}{#1}}

\newcommand{\Edit}[2]{[\textit{#1}$\to$\textit{#2}]}
\newcommand{\Decrease}[1]{\textcolor{blue}{$\Downarrow$ #1}}

\input{chapters/01_abstract}
\input{chapters/02_introduction}

\input{chapters/03_related_work}

\input{chapters/04_method}
\input{chapters/05_experiments}
\input{chapters/06_conclusion}

\bibliography{main}
\input{chapters/appendix}
\end{document}

%% file: chapters/01_abstract.tex
\begin{abstract}
The paper focuses on the interpretability of Grammatical Error Correction (GEC) evaluation metrics, which received little attention in previous studies. To bridge the gap, we introduce \textbf{\MetricName{}}, a reference-based metric describing four fundamental aspects of GEC systems: hit-correction, wrong-correction, under-correction, and over-correction. They collectively contribute to exposing critical qualities and locating drawbacks of GEC systems. Evaluating systems by combining these aspects also leads to superior human consistency over other reference-based and reference-less metrics. Extensive experiments on two human judgment datasets and six reference datasets demonstrate the effectiveness and robustness of our method, achieving a new state-of-the-art result. Our codes are released at \url{https://github.com/THUKElab/CLEME}.
\end{abstract}

%% file: chapters/02_introduction.tex
\section{Introduction}
The task of \textit{Grammatical Error Correction} (GEC) automatically detects and corrects grammatical errors in a given text~\cite{bryant2023grammatical}. A core component of GEC is the development of automatic metrics that can objectively measure model performance~\cite{kobayashi2024revisiting,ye-etal-2023-cleme}. However, proposing appropriate evaluation of GEC has long been a challenging task~\cite{madnani2011they}, due to the subjectivity~\cite{bryant-ng-2015-far}, complexity~\cite{mita2019cross} and subtlety~\cite{choshen2018inherent} of GEC.

\input{figures/intro}

Recent research efforts have focused on developing GEC metrics that closely align with human judgements~\cite{koyama-etal-2024-n-gram}, whereas the interpretability of these metrics has received less emphasis. We define the \textit{interpretability} of metrics as their \textit{capacity to disclose concerned characteristics of systems}, which is crucial for identifying weaknesses in a given GEC system. It is generally recognized that excellent systems should adhere to the principles of grammaticality and faithfulness~\cite{bryant2023grammatical}. Grammaticality demands that all grammatical errors be accurately corrected, while faithfulness ensures that corrections retain the original meaning and syntactic structure. Nevertheless, the commonly utilized GEC metrics~\cite{bryant-etal-2017-automatic,dahlmeier-ng-2012-better} are PRF-based (\textbf{P}recision, \textbf{R}ecall, and \textbf{F} scores). We claim that PRF-based metrics fail to effectively capture subtle dimensions of GEC systems, consequently hindering progress.
As illustrated in Figure~\ref{fig:intro}, the edits \Edit{were}{was} and \Edit{for}{in} in Hyp. 1 are regarded as 2 FP + FN edits by ERRANT~\cite{bryant-etal-2017-automatic} or 2 FP edits by CLEME~\cite{ye-etal-2023-cleme}. Meanwhile, the edit \Edit{$\epsilon$}{of} in Hyp. 2 is categorized as an FP edit for both ERRANT and CLEME. Despite this, these two categories of FP edits carry different implications. The former type is correctly placed but wrongly modified, whereas the latter is incorrectly positioned. The inability to differentiate between these FP edits results in ambiguous interpretations of P/R/F$_{0.5}$ scores, as they fail to quantify grammaticality and faithfulness.


Thus, we introduce \textbf{\MetricName{}}, an interpretable reference-based metric describing four fundamental aspects of GEC systems: 1) the \textit{hit-correction} score reflects the degree to which a system accurately corrects grammatical errors, 2) the \textit{wrong-correction} score denotes the degree of incorrect corrections made, 3) the \textit{under-correction} score reveals the degree of missing corrections, and 4) the \textit{over-correction} score measures the degree of excessive corrections. An excellent GEC system should gain a higher hit-correction score and lower wrong-correction, under-correction, and over-correction scores. The initial three aspects assess grammaticality, whereas the over-correction score pertains to faithfulness, given that it often alters the original meaning, a challenge notably observed with LLMs~\cite{coyne2023analyzing,DBLP:journals/corr/abs-2307-09007}. To achieve this, \MetricName{} first distinguishes between necessary and unnecessary \textit{false positive} (FP) edits. The idea is that necessary FP edits indicate the system's wrong-correction degree, while unnecessary FP edits reveal the system's over-correction degree. As shown in the bottom block of Figure~\ref{fig:intro}, \Edit{were}{was} and \Edit{for}{in} in Hyp. 1 are regarded as FP$_{\text{ne}}$ edits, while \Edit{$\epsilon$}{of} in Hyp. 2 is considered as an FP$_{\text{un}}$ edit.

As a result, \MetricName{} establishes a one-to-one relationship between four distinct system aspects and four types of edits: hit-correction v.s. TP, wrong-correction v.s. FP$_{\text{ne}}$, under-correction v.s. FN, and over-correction v.s. FP$_{\text{un}}$. Unlike conventional GEC metrics like ERRANT~\cite{bryant-etal-2017-automatic} and MaxMatch~\cite{dahlmeier-ng-2012-better} that evaluate using P/R/F$_{0.5}$ scores, it offers a nuanced view into the detailed aspects necessary for characterizing critical features of GEC systems. These separated scores are then consolidated into an overall score via linear weighted summation, giving varying importance to these distinct scores. This aggregate score provides a holistic measure of system performance. Similar to CLEME, our method adopts the chunk partition technique and supports evaluations based on either correction dependence or correction independence assumptions, so we dub the metric as \MetricName{}.


Moreover, we propose that edits of varying modification levels should uniquely influence the evaluation outcomes. For example, corrections involving punctuation are often less significant than corrections of content words. Therefore, we integrate two edit weighting techniques into \MetricName{}, similarity-based weighting~\cite{gong-etal-2022-revisiting} and LLM-based weighting. In particular, these methods compute a specific weight for each edit through a language model rather than assigning equal weight to all edits, thereby enabling \MetricName{} to grasp contextual semantics and address the limitations of conventional metrics that depend on surface-level form similarity~\cite{kobayashi2024large}.

To verify the effectiveness of \MetricName{}, we conduct extensive experiments on two human judgment datasets (GJG15~\cite{grundkiewicz2015human} and SEEDA~\cite{kobayashi2024revisiting}), where our method consistently achieves high correlations. We also demonstrate the robustness of \MetricName{} by computing the evaluation results based on six reference datasets with disparate annotation styles. In summary, our contributions are three folds:

\begin{itemize}
    \item [(1)] We introduce \MetricName{}, an interpretable reference-based metric, which is beneficial to reveal crucial aspects of GEC systems.

    \item[(2)] We enhance \MetricName{} by incorporating two edit weighting techniques, addressing the limitations of conventional reference-based metrics in capturing semantics.

    \item[(3)] Extensive experiments and analyses are conducted to confirm the effectiveness and robustness of our proposed method.
\end{itemize}


%% file: figures/intro.tex
\begin{figure}[tbp!]
    \centering
    \includegraphics[width=\linewidth]{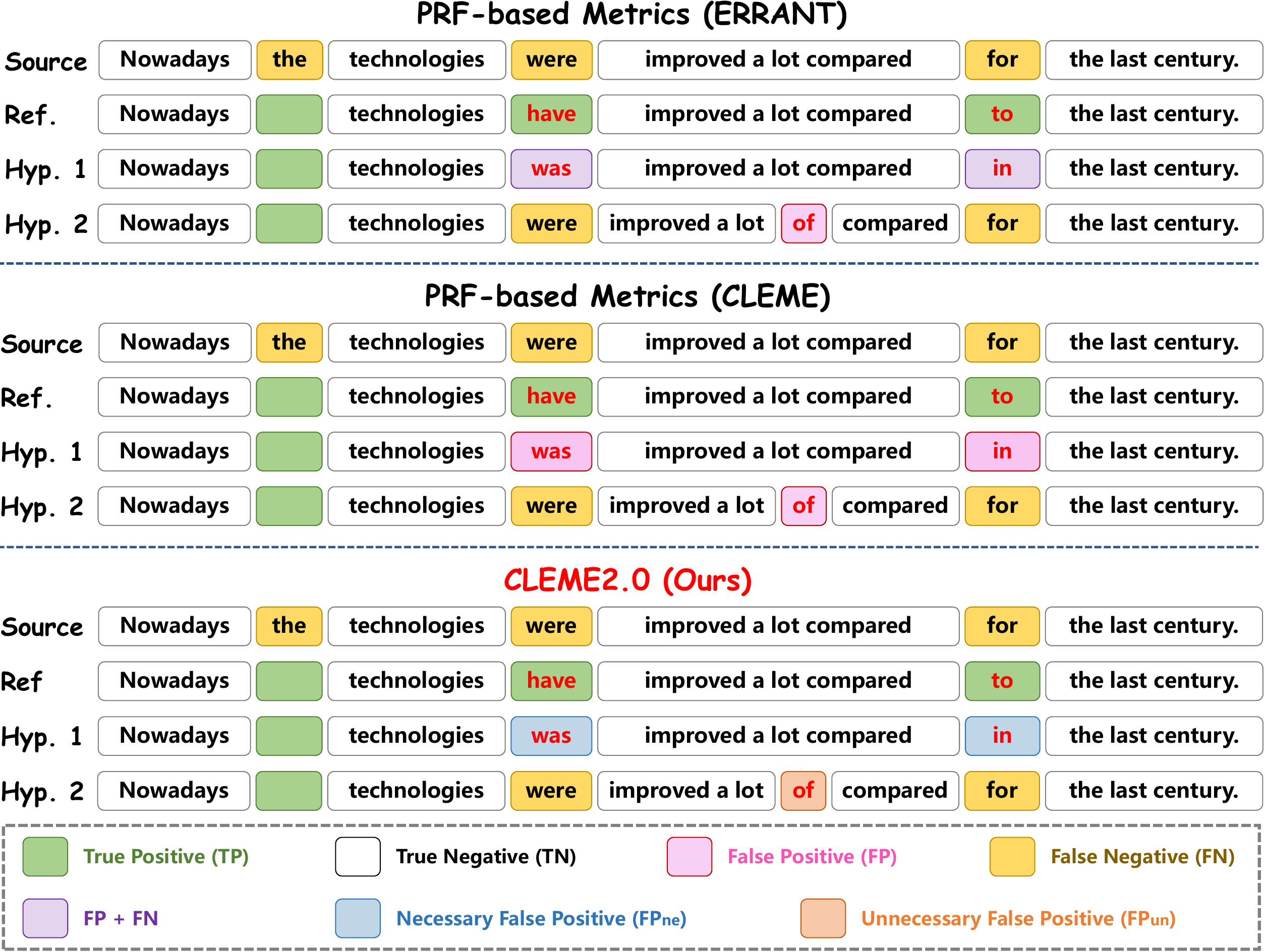}
    \caption{An example of \MetricName{}. We highlight \TP{TP}, \FP{FP}, \FPne{FP$_{\text{ne}}$}, \FPun{FP$_{\text{un}}$}, and \FN{FN} in different colors.}
    \label{fig:intro}
\end{figure}

%% file: chapters/03_related_work.tex
\input{figures/overview}

\section{Related Work}
\paragraph{Reference-based metrics.}

Reference-based metrics evaluate GEC systems by comparing their outputs to manually written references~\cite{ye2022focus,DBLP:conf/emnlp/YeLL023,ye2023system,DBLP:conf/emnlp/HuangYZLLZZ23, DBLP:journals/corr/abs-2402-11420,DBLP:conf/acl/LiZLLLSWLCZ22,DBLP:conf/emnlp/LiMZLLHLLC022,DBLP:conf/acl/LiXC0LMJLZZS24,DBLP:conf/emnlp/MaLSZHZLLLCZS22,DBLP:conf/icassp/ZhangLZMLCZ23,DBLP:journals/corr/abs-2501-00334,DBLP:journals/eswa/LiMCHHLZS25}. The M$^2$ scorer~\cite{dahlmeier2012better} identifies optimal edit sequences between source sentences and system hypotheses, using F$_{0.5}$ scores. However, this method can inflate scores by manipulating edit boundaries. To mitigate this problem, ERRANT~\cite{bryant-etal-2017-automatic} improves edit extraction through a linguistically-informed alignment algorithm, but it remains language-dependent and biased in multi-reference evaluation. CLEME~\cite{ye-etal-2023-cleme} further provides unbiased F$_{0.5}$ scores and introduces an extra correction assumption for multi-reference evaluation. PT-M$^2$~\cite{gong-etal-2022-revisiting} combines PT-based and existing GEC metrics for higher correlations with human judgments.

\paragraph{Reference-less metrics.}

To overcome the limitations of reference-based metrics, recent studies focus on reference-less scoring. Inspired by quality estimation in NMT~\cite{DBLP:journals/patterns/LiuLTLZ22, DBLP:journals/csur/DongLGCLSY23}, ~\citet{napoles2016there} propose Grammaticality-Based Metrics (GBMs) using an existing GEC system or a pre-trained ridge regression model.~\citet{asano2017reference} enhance GBMs by adding criteria like grammaticality, fluency, and meaning preservation. \citet{yoshimura2020some} introduce SOME, which uses sub-metrics optimized for manual assessment with regression models. Scribendi Score~\cite{islam2021end} combines language perplexity and token/Levenshtein distance ratios. IMPARA~\cite{maeda2022impara} incorporates a Quality Estimator and a Semantic Estimator based on BERT to evaluate GEC output quality and semantic similarity. While reference-less metrics align well with human judgments, they lack interpretability due to the heavy dependence on trained models, thus posing latent risks.

%% file: figures/overview.tex
\begin{figure*}[tb!]
\centering
\includegraphics[scale=0.30]{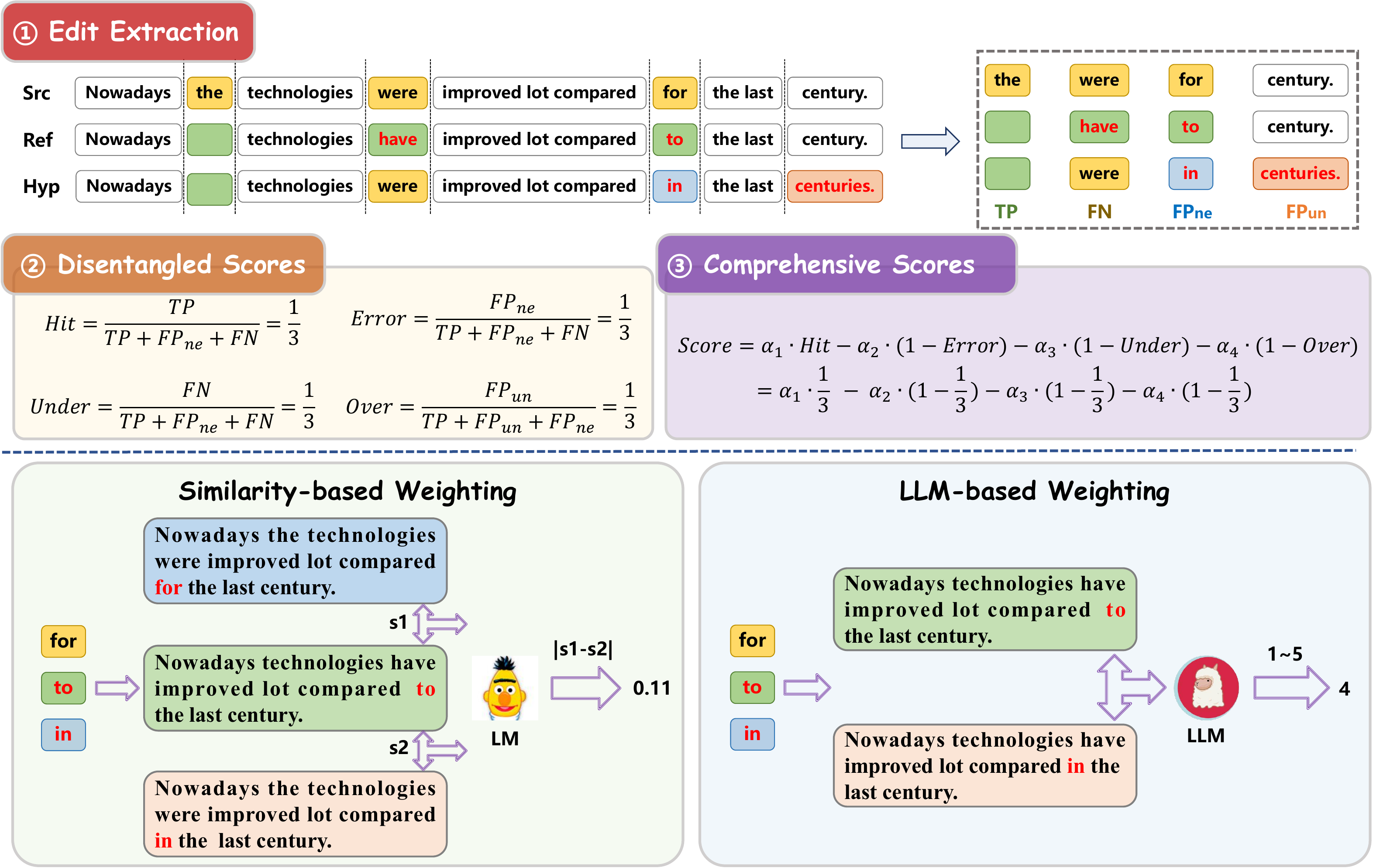}
\caption{
Overview of \MetricName{}. Initially, we extract edits and categorize hypothesis edits as \TP{TP}, \FN{FN}, \FPne{FP$_{\text{ne}}$}, and \FPun{FP$_{\text{un}}$}. Next, we compute four distinct scores. Finally, we integrate these scores into an overall score utilizing one of the edit weighting techniques.
}
\label{fig:overview}
\end{figure*}

%% file: chapters/04_method.tex
\section{Method}
Our \MetricName{} can be generally divided into three main steps, with the overview shown in Figure~\ref{fig:overview}. Additionally, we incorporate two distinct edit weighting techniques to enhance performance.

\subsection{Edit Extraction}
Given a source sentence $X$ and a target (either hypothesis or reference) sentence $Y$, we extract the edits describing the modification from $X$ to $Y$. Here, we utilize the chunk partition technique from CLEME~\cite{ye-etal-2023-cleme} to execute the process of edit extraction. Unlike the traditional metrics like ERRANT~\cite{bryant-etal-2017-automatic} and MaxMatch~\cite{dahlmeier-ng-2012-better}, CLEME concurrently aligns all sentences, including the source, the hypothesis, and all the references. This facilitates segmentation of them all into chunk sequences with an equal number of chunks, irrespective of the varying token counts in different sentences, as delineated in Figure~\ref{fig:overview}. It is worth noting that a chunk is a basic edit unit, which can be unchanged, corrected, or dummy (empty)~\cite{ye-etal-2023-cleme}.

\subsection{Disentangled Scores}
To compute disentangled scores, we initially disentangle edits into four elementary types. 
1) \textbf{TP edits} refer to the corrected/dummy hypothesis chunks that share the same tokens as the corresponding reference chunks. 
2) \textbf{FP$_{\text{ne}}$ edits} are the corrected/dummy hypothesis chunks that have different tokens from those in the corresponding reference chunks wherein the reference chunks are also corrected/dummy ones. 
3) \textbf{FP$_{\text{un}}$ edits} are the corrected hypothesis chunks but their corresponding reference chunks remain unchanged. 
4) \textbf{FN edits} indicate the unchanged hypothesis chunks but the corresponding reference chunks are corrected/dummy.
It is highlighted that traditional metrics~\cite{dahlmeier-ng-2012-better,bryant-etal-2017-automatic,DBLP:journals/corr/abs-2311-11268} do not distinguish between FP$_{\text{ne}}$ and FP$_{\text{un}}$, treating both as FP, thereby resulting in confusion between wrong-correction and over-correction. Actually, we have $FP = FP_{\text{ne}} + FP_{\text{un}}$.

Furthermore, we can differentiate between necessary and unnecessary edits. TP, FP$_{\text{ne}}$, and FN edits are all \textit{necessary} edits, since their corresponding reference chunks are also corrected/dummy, implying the existence of grammatical errors in the related parts of $X$. On the contrary, FP$_{\text{un}}$ edit are \textit{unnecessary} edits because the systems propose corrections not represented in references. Consequently, we can define four disentangled scores.

\paragraph{Hit-correction score.} This paper defines the hit-correction score as the ratio of TP edits to all necessary reference edits. Its purpose is to quantify the accuracy with which systems offer correct corrections. The formula is as follows:

\begin{small}\begin{equation}
Hit = \frac{TP}{necessity} = \frac{TP}{TP+FP_{\text{ne}}+FN}
\label{eq:hit}\end{equation}\end{small}

\paragraph{Wrong-correction score.} Conversely, the wrong-correction score is defined as the ratio of FP$_{\text{ne}}$ edits to all necessary reference edits. This score seeks to evaluate the degree to which systems generate erroneous corrections for grammatical errors. The formula for this score is as follows:

\begin{small}\begin{equation}
Wrong = \frac{FP_{\text{ne}}}{necessity} = \frac{FP_{\text{ne}}}{TP+FP_{\text{ne}}+FN}
\end{equation}\end{small}

\paragraph{Under-correction score.} Similarly, the under-correction score is proposed to measure the degree to which systems omit to correct grammatical errors, which is computed as follows:

\begin{small}\begin{equation}
Under = \frac{FN}{necessity} = \frac{FN}{TP+FP_{\text{ne}}+FN}
\end{equation}\end{small}


\paragraph{Over-correction score.} The score is introduced in response to frequent observations that LLMs are prone to over-correcting texts. This score is determined by the proportion of FP$_{\text{un}}$ edits to all hypothesis corrected/dummy edits, aiming to gauge the level to which systems offer excessive corrections:

\begin{small}\begin{equation}
Over = \frac{FP_{\text{un}}}{TP+FP} = \frac{FP_{\text{un}}}{TP+FP_{ne}+FP_{un}}
\label{eq:over}\end{equation}\end{small}

With the disentangled scores indicating disparate aspects of GEC systems, researchers can identify specific weaknesses and implement targeted improvements without expensive human labor.

\subsection{Comprehensive Score}
\label{subsec:comprehensive_score}
Once the four disentangled scores have been computed, they need to be merged into a comprehensive score that encapsulates the global performance of the systems. We employ a weighted summation approach to organize these four scores for interpretability and simplification. By definition, systems with higher hit-correction scores are usually preferable, a tendency that inversely applies to the remaining scores. Thus, the comprehensive score can be calculated using the following formula:

\begin{small}\begin{equation}\begin{aligned}
& Score = \alpha_{1}\cdot Hit + \alpha_{2}\cdot (1-Wrong) \\
& ~~ + \alpha_{3}\cdot (1-Under) + \alpha_{4}\cdot (1-Over)
\end{aligned}\end{equation}\end{small}

\noindent where $\alpha_i$ is the trade-off factor for each disentangled score, and we constrain that $0 < \alpha_i < 1$ and $\alpha _{1}+ \alpha _{2}+ \alpha _{3}+ \alpha _{4} = 1$. 

\subsection{Edit Weighting}
Existing reference-based metrics, such as ERRANT and CLEME, depend heavily on superficial literal similarity. This means that, regardless of length or modification, all types of edits have equal weighting in the evaluation scores. This aspect fails to acknowledge that human evaluators might semantically consider the edits' varying importance levels. Therefore, we introduce two distinct edit weighting techniques to compute the importance weights of edits. These weights are then incorporated into the calculation of the aforementioned disentangled scores as depicted in Equation~\eqref{eq:hit} $\sim$ \eqref{eq:over}. Take the hit-correction score as a typical example, we reformulate the Equation~\eqref{eq:hit} as follows:

\begin{small}\begin{equation}
Hit = \frac{w_{TP}}{w_{TP} + w_{FP_{\text{ne}}} + w_{FN}}
\end{equation}\end{small}

\paragraph{Similarity-based weighting.}
We use PTScore to assign edit weights~\cite{gong-etal-2022-revisiting}. Since it performs based on BERTScore~\cite{zhang2019bertscore}, a tool for evaluating text generation through similarity scores, we refer to this technique as similarity-based weighting. The rationale is to prioritize edits with a more significant modification of the meaning and quality of the text.

By simulating a partially accurate version $X'$ of the source sentence $X$, PTScore can associate specific weights to edits within a sentence. The computation process is as follows:

\begin{small}\begin{equation}
X' = \operatorname{replace}(X, e_{\text{hyp}})
\end{equation}\end{small}
\begin{small}\begin{equation}
w = \operatorname{PTScore}(X', R) - \operatorname{PTScore}(X, R)
\end{equation}\end{small}

\noindent where $R$ is the reference sentence, while the function $\operatorname{replace}()$ is used to replace a specific chunk of the source $X$ with the corrected/dummy hypothesis chunk $e_{\text{hyp}}$. A positive weight $w > 0$ indicates a beneficial correction, whereas a negative value suggests a wrong correction. The absolute value $|w|$ is utilized as the edit weight following~\citep{gong-etal-2022-revisiting}, and the significance\footnote{For more detailed analysis, refer to our case study in Section~\ref{subsec:case_study} and PT-M$^2$~\cite{gong-etal-2022-revisiting}.} of an edit in a sentence grows with a larger $|w|$.

\paragraph{LLM-based weighting.}
Recent studies have begun investigating the effectiveness of LLM-based evaluation, known for their advanced semantic comprehension~\cite{DBLP:journals/corr/abs-2404-04925}, in assessing various NLP tasks~\cite{pavlovic2024effectiveness,sottana2023evaluation}. Building on this trend, we prompt Llama-2-7B~\cite{touvron2023llama} to assign edit weights from 1 to 5, where a higher value signifies more critical edits. This methodology is rooted in the idea that LLMs, due to their extensive training on diverse data, are adept at grasping intricate language patterns and text structure. Detailed implementation instructions and the prompting framework are available in Appendix~\ref{appendix:llm}.

%% file: chapters/05_experiments.tex
\section{Experiments}

\subsection{Experimental Settings}

\input{tables/GJG15_all}


\paragraph{Human ranking datasets.}
We conduct comprehensive experiments across two human judgment datasets with disparate annotation protocols.

\begin{itemize}
    \item \textbf{GJG15}~\cite{grundkiewicz2015human} is constructed to manually evaluate classical systems~\cite{junczys2014amu,rozovskaya2014illinois} in the CoNLL-2014 shared task~\cite{ng2014conll}.

    \item \textbf{SEEDA}. \citet{kobayashi2024revisiting} reveal several shortcomings in GJS15 and subsequently propose SEEDA, an alternative dataset featuring human judgments across two levels of granularity. To align with the contemporary trend in GEC, SEEDA is primarily focused on mainstream neural-based systems.
\end{itemize}

Both of human judgment datasets derive the overall human rankings for all GEC systems by employing Expected Wins (EW)~\cite{bojar-etal-2013-findings} and TrueSkill (TS)~\cite{sakaguchi2014efficient} methods. Following the previous approaches~\cite{ye-etal-2023-cleme,kobayashi2024revisiting}, we compute the Pearson ($\gamma$) and Spearman ($\rho$) correlations between metrics and human judgments, in order to ascertain the effectiveness and robustness of GEC metrics within the context of \textit{system-level ranking}.

\paragraph{Reference datasets.}
Reference-based metrics rely on a reference set to establish a system ranking list, the properties of which may significantly influence the performance of the metrics. To investigate the impact of variable reference sets, we assess human consistency across 6 reference datasets. These datasets encompass a range of annotation styles, and a number of human annotators, including CoNLL-2014~\cite{grundkiewicz2015human}, BN-10GEC~\cite{bryant-ng-2015-far} and SN-8GEC~\cite{sakaguchi-etal-2016-reassessing}. Notably, SN-8GEC is partitioned into 4 sub-sets, i.e., \underline{E}xpert-Minimal, \underline{E}xpert-Fluency, \underline{N}on-\underline{E}xpert-Minimal, and \underline{N}on-\underline{E}xpert-Fluency. A more thorough breakdown of these datasets and the statistics is provided in Appendix~\ref{appendix:GEC_Meta_Evaluation}.

\paragraph{Corpus and sentence levels.}
GEC evaluation metrics can compute an overall system-level score for a given system in two settings~\cite{gong-etal-2022-revisiting}. Given the metric $M$, source sentences $\mathbf{S}$, hypothesis sentences $\mathbf{H}$ and reference sentences $\mathbf{R}$, 1) \textbf{corpus-level} metrics compute the system score based on the whole corpus $M(\mathbf{S},\mathbf{H},\mathbf{R})$, and 2) \textbf{sentence-level} metrics use the average of the sentence-level scores $ {\textstyle \sum_{i}^{I}} M(\mathbf{S}_{i}, \mathbf{H}_{i}, \mathbf{R}_{i})/I$. 

\paragraph{Trade-off factors.}
We leverage a cross-evaluation search method to identify two optimal sets of trade-off factors for both the corpus and sentence levels. At the corpus level, the factors are assigned as $\alpha_{1}, \alpha_{2}, \alpha_{3}, \alpha_{4} = 0.45, 0.35, 0.15, 0.05$, while for the sentence level, they are adjusted to $\alpha_{1}, \alpha_{2}, \alpha_{3}, \alpha_{4} = 0.35, 0.25, 0.20, 0.20$. The details of the chosen values of trade-off factors can be seen in Appendix~\ref{appendix:trade_off}.

\paragraph{Evaluation assumptions.}
CLEME can evaluate GEC systems based on correction \textit{dependence} (-dep) or \textit{independence} (-ind) assumptions. The correction independence assumption offers a more relaxed edit-matching process, implying that systems might yield better scores when multiple references are available. Inspired by this work, \MetricName{} also supports both assumptions, and we will study their effects on our method.

\subsection{Results of GJG15 Ranking}
The correlations between the GEC metrics and human judgments on the GJG15 rankings are shown in Table~\ref{tab:exp-main-all}, and we have the following insights.

\paragraph{\MetricName{} outperforms other metrics at both corpus and sentence levels.} For corpus-level, \MetricName{}-sim-ind achieves the highest average correlations, closely followed by \MetricName{}-sim-dep. \MetricName{}-ind and \MetricName{}-dep can also gain comparable correlations with other metrics, even though they do not utilize any edit weighting techniques. On the other hand, sentence-level metrics exhibit a similar pattern. Sent\MetricName{}-sim-dep and Sent\MetricName{}-sim-ind achieve the highest Pearson and Spearson correlations, respectively. These results significantly demonstrate the effectiveness and robustness of our proposed method across different settings.

\paragraph{Sentence-level metrics outperform their corpus-level counterparts.}
This observation is consistent with recent studies~\cite{gong-etal-2022-revisiting,ye-etal-2023-cleme}. This is because system-level rankings treat each sample equally regardless of edit numbers, mirroring how sentence-level metrics are evaluated. On the other hand, corpus-level metrics emphasize samples with more edits, thus causing the gap between automatic metrics and human evaluation. SentPT-M$^2$ shows superior performance on CoNLL-2014 but performs worse on BN-10GEC, E-Minimal, and NE-Fluency compared to our approach, revealing a lack of robustness of the metric.

Generally, our method aligns more closely with human assessments than existing popular metrics. Notably, our method with similarity-based weighting surpasses unweighted ones, thanks to the integration of semantic factors. However, on E-Minimal and NE-Minimal, weighted and unweighted results are comparable. We suspect this is because these datasets have minimal yet crucial annotations, reducing the possibility of varying weights and the efficacy of edit weighting.

Furthermore, we present comprehensive results of \MetricName{} on CoNLL-2014 and offer insights into our method for analyzing and identifying weaknesses in GEC systems in Appendix~\ref{appendix:detailed_results}.

\input{tables/SEEDA-base}

\subsection{Results of SEEDA Ranking}
We carry out an additional experiment on the SEEDA-Sentence and SEEDA-Edit datasets, where we compare our method against various GEC metrics. As presented in Table \ref{tab:seeda-base}, our approach consistently achieves the best outcomes across both datasets. According to~\citet{kobayashi2024revisiting}, the correlations of most metrics tend to decrease when transitioning from classical to neural evaluation systems. This implies that conventional metrics might face difficulties in evaluating the more extensively edited or fluent corrections produced by state-of-the-art neural GEC systems. Nevertheless, our method effectively tackles these challenges, delivering even improved performance for all indicators. The results for SEEDA-Edit exceed those for SEEDA-Sentence, due to the greater detail in SEEDA-Edit, aligning more closely with the operation of \MetricName{}.

It is crucial to mention that reference-less metrics such as SOME and IMPARA yield high outcomes, in part, because these are fine-tuned on GEC data. Although fine-tuned metrics generally perform better, they are not without their limitations. Firstly, the incorporation of fine-tuning in SOME and IMPARA makes these reference-less metrics more costly. Second, these reference-less metrics may suffer from poor robustness since the assessment process is not guided by human-annotated references. For example, the authors of Scribendi Score claim that it can achieve high correlations on the human judgment dataset from~\citet{napoles-etal-2016-theres}. However, only moderate correlations are observable on SEEDA-Edit.

\input{tables/case-study}

\input{tables/LLM_results}

\subsection{Results of LLM-based Weighting}
\label{subsec:llm}
Table~\ref{tab:llm} presents the outcomes of LLM-based weighting, noting that its effectiveness is less favorable than similarity-based weighting. A likely reason is the coarse grading method of LLMs, which allocates edit weights from 1 to 5, unlike the finer continuous scale $[0,1]$. Although \citet{kobayashi2024large} argue that LLMs serve as effective evaluators for GEC, their research pertains to huge closed-source LLMs (GPT-4 and GPT-3.5) and involves specific prompt engineering. They also identify the importance of the LLM scale since GPT-3.5 may even obtain negative correlations with human judgments. In contrast, we employ a more straightforward approach with open-source LLama-2-7B.

\section{Analysis}
\label{app:analysis}

\subsection{Case Study}
\label{subsec:case_study}
Table~\ref{tab:case-study} demonstrates instances of \MetricName{}. In the first set, Chunks 3 and 5 are FP$_{ne}$ edits contributing to the wrong-correction score, with a higher edit weight of Chunk 5 than Chunk 3 since Chunk 5 introduces an error that entirely alters the sentence's meaning.
In the second set, Chunk 2 obtains the highest edit weight of 0.056, underscoring its substantial influence on the evaluation. Despite the correct modification of ``diagnosed'', the misuse of ``out'' remains, keeping the correction wrong. Chunk 4 illustrates a singular-to-plural correction in the source sentence, with a low weight indicating a minor impact. Chunks 6 and 8 showcase over-corrections. Chunk 6 leaves the original meaning unchanged, whereas Chunk 8 introduces a significant error by misusing a personal pronoun.

The cases highlight the effectiveness of the weighting technique. Otherwise, all edits are given equal weight, failing to distinguish hypothesis edits with varying correction levels. We display the cases of LLM-based weighting in Appendix~\ref{app:case_llm}.

\subsection{Ablation Study}
\label{subsec:ablation}
\input{tables/ablation_disentangled_score}

We conduct ablation studies on (Sent)\MetricName{}-dep to analyze the performance of individual disentangled scores. A preferable system has reduced wrong-correction, under-correction, and over-correction scores, so we report corrections between 1-$x$ with human judgments where $x$ is one of the scores. The outcomes are detailed in Table \ref{tab:disentangled_score}. Hit-correction and under-correction show moderate correlations. Over-correction scores have small positive correlations at the corpus level, with minimal negative correlations at the sentence level. Notably, wrong-correction scores display negative correlations, but this does not mean they do not impact the overall score. In reality, the trade-off factor for wrong-correction scores is relatively substantial. The hypothesis is that focusing evaluations only on wrong-correction scores might prefer systems that make only highly confident edits, potentially leading to assessment bias.

Additionally, we utilize the similarity-weighting approach on CLEME to evaluate its efficacy, with the outcomes detailed in Appendix~\ref{appendix:extra_results_cleme}. To examine our method on a broad scale, we also provide the average correlations obtained from a comprehensive analysis of all potential parameter settings. The results are found in Appendix~\ref{appendix:average_correlations}.

\input{tables/efficiency}

\subsection{Efficiency}
This section provides a comparative analysis of the efficiency of our methods against other prevailing metrics. The experiments were executed on a GPU 3090 within the CoNLL-2014 framework, with the evaluation times of the AMU system reported. Our observations are as follows: (1) For ERRANT, the primary time expenditure is associated with edit extraction, lasting 33.4 seconds. (2) CLEME and \MetricName{} primarily incur time costs from edit extraction at 33.4 seconds and chunk partitioning at 20.7 seconds. (3) For \MetricName{}-sim, the most significant time costs are assignable to edit extraction (33.4 s), chunk partitioning (20.7 s), and edit weighting (34.3 s). PT-M2 exhibits the slowest runtime when replicating existing mainstream methods, with its evaluation process taking several hours; thus, we did not report a precise runtime due to the time constraints. Some technical solutions can mitigate the runtime when evaluating a system using these metrics concurrently. For instance, when assessing a system with ERRANT, CLEME, and \MetricName{}, the minimum cumulative duration is calculated as 33.4 seconds for edit extraction, 20.7 seconds for chunk partitioning, and 34.3 seconds for edit weighting, totaling 88.4 seconds.

%% file: tables/GJG15_all.tex
\begin{table*}[tb!]
\renewcommand{\arraystretch}{1.2}
\renewcommand{\tabcolsep}{4pt}
\resizebox{1.0\linewidth}{!}{
\begin{tabular}{lclllllllllllll}
\toprule

\multirow{2.5}{*}{\textbf{Metric}} & &
\multicolumn{2}{c}{\textbf{CoNLL-2014}} &
\multicolumn{2}{c}{\textbf{BN-10GEC}} &
\multicolumn{2}{c}{\textbf{E-Minimal}} & 
\multicolumn{2}{c}{\textbf{E-Fluency}} &
\multicolumn{2}{c}{\textbf{NE-Minimal}} &
\multicolumn{2}{c}{\textbf{NE-Fluency}} &
\multicolumn{1}{c}{\multirow{2}{*}{\textbf{Avg.}}} \\

\cmidrule(lr){3-4} \cmidrule(lr){5-6} \cmidrule(lr){7-8} \cmidrule(lr){9-10} \cmidrule(lr){11-12} \cmidrule(lr){13-14}

& & \multicolumn{1}{c}{\textbf{EW}} & \multicolumn{1}{c}{\textbf{TS}}
& \multicolumn{1}{c}{\textbf{EW}} & \multicolumn{1}{c}{\textbf{TS}}
& \multicolumn{1}{c}{\textbf{EW}} & \multicolumn{1}{c}{\textbf{TS}}
& \multicolumn{1}{c}{\textbf{EW}} & \multicolumn{1}{c}{\textbf{TS}}
& \multicolumn{1}{c}{\textbf{EW}} & \multicolumn{1}{c}{\textbf{TS}}
& \multicolumn{1}{c}{\textbf{EW}} & \multicolumn{1}{c}{\textbf{TS}}
\\

\midrule

\multirow{2}{*}{\textbf{M$^2$}}
& $\gamma$ & 0.623 & 0.672 & 0.547 & 0.610 & 0.597 & 0.650 & 0.590 & 0.659 & 0.575 & 0.634 & 0.582 & 0.649 & 0.616\\
& $\rho$   & 0.687 & 0.720 & 0.648 & 0.692 & 0.654 & 0.703 & 0.654 & 0.709 & 0.577 & 0.648 & 0.648 & 0.703 & 0.670 \\

\hdashline

\multirow{2}{*}{\textbf{GLEU}}
& $\gamma$ & 0.701 & 0.750 & 0.678 & 0.761 & 0.533 & 0.513 & 0.693 & 0.771 & -0.044 & -0.113 & 0.674 & 0.767 & 0.557 \\
& $\rho$   & 0.467 & 0.555 & 0.754 & 0.806 & 0.577 & 0.511 & 0.710 & 0.757 & -0.005 & -0.055 & 0.725 & 0.819 & 0.551 \\

\hdashline

\multirow{2}{*}{\textbf{ERRANT}}
& $\gamma$ & 0.642 & 0.688 & 0.586 & 0.644 & 0.578 & 0.631 & 0.594 & 0.663 & 0.585 & 0.637 & 0.597 & 0.659 & 0.625 \\
& $\rho$   & 0.659 & 0.698 & 0.637 & 0.698 & 0.742 & 0.786 & 0.720 & 0.775 & 0.747 & 0.797 & 0.753 & 0.797 & 0.734\\

\hdashline

\multirow{2}{*}{\textbf{PT-M$^2$}}
& $\gamma$ & 0.693 & 0.737 & 0.650 & 0.706 & 0.626 & 0.667 & 0.621 & 0.681 & 0.630 & 0.675 & 0.620 & 0.682 & 0.666 \\
& $\rho$   & 0.758 & 0.769 & 0.690 & 0.824 & 0.709 & 0.736 & 0.758 & 0.802 & 0.736 & 0.758 & 0.758 & 0.802 & 0.758 \\

\hdashline



\multirow{2}{*}{\textbf{CLEME-dep} }
& $\gamma$ & 0.648 & 0.691 & 0.602 & 0.656 & 0.594 & 0.644 & 0.589 & 0.654 & 0.595 & 0.643 & 0.612 & 0.673 & 0.633 \\
& $\rho$   & 0.709 & 0.742 & 0.692 & 0.747 & \underline{0.797} & 0.813 & 0.714 & 0.775 & 0.786 & 0.835 & 0.720 & 0.791 & 0.760 \\

\hdashline

\multirow{2}{*}{\textbf{CLEME-ind}}
& $\gamma$ & 0.649 & 0.691 & 0.609 & 0.659 & 0.593 & 0.643 & 0.587 & 0.653 & 0.601 & 0.647 & 0.611 & 0.672 & 0.635 \\
& $\rho$   & 0.709 & 0.731 & 0.692 & 0.747 & 0.791 & 0.802 & 0.731 & 0.791 & 0.797 & 0.841 & 0.714 & 0.786 & 0.761 \\

\hdashline

\multirow{2}{*}{\textbf{\MetricName{}-dep} (Ours)}
& $\gamma$ & 0.700 & 0.765 & 0.675 & 0.745 & 0.690 & 0.768 & 0.695 & 0.788 & 0.702 & 0.778 & 0.704 & 0.800 & 0.734 \\
& $\rho$   & 0.665 & 0.736 & 0.626 & 0.692 & 0.736 & 0.808 & 0.742 & 0.830 & 0.775 & 0.846 & 0.599 & 0.714 & 0.730 \\

\hdashline

\multirow{2}{*}{\textbf{\MetricName{}-ind} (Ours)}
& $\gamma$ & 0.718 & 0.777 & \underline{0.731} & 0.793 & 0.708 & 0.784 & 0.736 & 0.824 & 0.757 & \underline{0.826} & \underline{0.801} & \underline{0.848} & 0.775 \\
& $\rho$   & 0.665 & 0.736 & 0.698 & 0.758 & 0.736 & 0.808 & 0.742 & 0.830 & 0.775 & 0.846 & 0.670 & 0.769 & 0.753 \\

\hdashline

\multirow{2}{*}{\textbf{\MetricName{}-sim-dep} (Ours)}
& $\gamma$ & \underline{0.783} & \underline{0.853} & 0.721 & \underline{0.801} & \underline{0.765} & \underline{0.834} & \underline{0.737} & \underline{0.827} & \underline{0.761} & 0.824 & 0.741 & 0.834 & \underline{0.790} \\
& $\rho$   & \underline{0.819} & \underline{0.890} & \underline{0.802} & \underline{0.863} & 0.791 & \underline{0.868} & \underline{0.758} & \underline{0.852} & \underline{0.830} & \underline{0.896} & \underline{0.786} & \underline{0.857} & \underline{0.834} \\

\hdashline

\multirow{2}{*}{\textbf{\MetricName{}-sim-ind} (Ours)}
& $\gamma$ & \bf{0.806} & \bf{0.871} & \bf{0.772} & \bf{0.839} & \bf{0.780} & \bf{0.841} & \bf{0.761} & \bf{0.844} & \bf{0.782} & \bf{0.834} & \bf{0.798} & \bf{0.877} & \bf{0.817} \\
& $\rho$   & \bf{0.846} & \bf{0.901} & \bf{0.835} & \bf{0.885} & \bf{0.819} & \bf{0.885} & \bf{0.758} & \bf{0.852} & \bf{0.846} & \bf{0.896} & \bf{0.863} & \bf{0.923} & \bf{0.859} \\



\hline\hline

\multirow{2}{*}{\textbf{SentM$^2$}}
& $\gamma$ & 0.871 & 0.864 & 0.567 & 0.646 & 0.805$^\clubsuit$ & 0.836$^\clubsuit$ & 0.655 & 0.732 & 0.729$^\clubsuit$ & 0.785$^\clubsuit$ & 0.621 & 0.699  & 0.734 \\
& $\rho$   & 0.731 & 0.758 & 0.593 & 0.648 & 0.806$^\clubsuit$ & 0.845$^\clubsuit$ & 0.731 & 0.764 & 0.797$^\clubsuit$ & 0.846$^\clubsuit$ & 0.632 & 0.687 & 0.737 \\

\hdashline

\multirow{2}{*}{\textbf{SentGLEU}}
& $\gamma$ & 0.784 & 0.828 & 0.756 & 0.826 & 0.742$^\clubsuit$ & 0.773$^\clubsuit$ & 0.785 & 0.846 & 0.723$^\clubsuit$ & 0.762$^\clubsuit$ & 0.778 & 0.848 & 0.788 \\
& $\rho$   & 0.720 & 0.775 & 0.769 & 0.824 & 0.764$^\clubsuit$ & 0.797$^\clubsuit$ & 0.791 & 0.846 & 0.764$^\clubsuit$ & 0.830$^\clubsuit$ & 0.768 & 0.846 & 0.791 \\

\hdashline

\multirow{2}{*}{\textbf{SentERRANT} }
& $\gamma$ & 0.870 & 0.846 & \underline{0.885} & \underline{0.896} & 0.768$^\clubsuit$ & 0.803$^\clubsuit$ & 0.806 & 0.732 & 0.710$^\clubsuit$ & 0.765$^\clubsuit$ & 0.793 & 0.847 & 0.810 \\
& $\rho$   & 0.742 & 0.747 & 0.786 & 0.830 & 0.775$^\clubsuit$ & 0.819$^\clubsuit$ & 0.813 & 0.764 & 0.780$^\clubsuit$ & 0.841$^\clubsuit$ & 0.830 & 0.857 & 0.799 \\

\hdashline

\multirow{2}{*}{\textbf{SentPT-M$^2$}}
& $\gamma$ & \bf{0.949} & \bf{0.938} & 0.602$^\clubsuit$ & 0.682$^\clubsuit$ & 0.831$^\clubsuit$ & 0.855$^\clubsuit$ & 0.689 & 0.763 & 0.770$^\clubsuit$ & 0.822$^\clubsuit$ & 0.648 & 0.725 & 0.772 \\
& $\rho$   & \underline{0.907} & 0.874 & 0.626$^\clubsuit$ & 0.670$^\clubsuit$ & 0.808$^\clubsuit$ & 0.819$^\clubsuit$ & 0.797 & 0.841 & 0.813$^\clubsuit$ & 0.857$^\clubsuit$ & 0.742 & 0.786 & 0.795 \\

\hdashline



\multirow{2}{*}{\textbf{SentCLEME-dep}}
& $\gamma$ & 0.876 & 0.844 & \bf{0.915} & \bf{0.913} & 0.806$^\clubsuit$ & 0.838$^\clubsuit$ & 0.849 & 0.886 & 0.742$^\clubsuit$ & 0.795$^\clubsuit$ & 0.876 & 0.921 & 0.855 \\
& $\rho$   & 0.824 & 0.808 & \bf{0.835} & \bf{0.874} & 0.775$^\clubsuit$ & 0.819$^\clubsuit$ & 0.824 & 0.863 & 0.797$^\clubsuit$ & 0.846$^\clubsuit$ & 0.791 & 0.846 & 0.825 \\

\hdashline

\multirow{2}{*}{\textbf{SentCLEME-ind} }
& $\gamma$ & 0.868 & 0.857 & 0.855$^\clubsuit$ & 0.876$^\clubsuit$ & 0.821$^\clubsuit$ & 0.856$^\clubsuit$ & 0.841 & 0.877 & 0.782$^\clubsuit$ & 0.831$^\clubsuit$ & 0.852 & 0.896 & 0.851 \\
& $\rho$   & 0.725 & 0.758 & 0.659$^\clubsuit$ & 0.714$^\clubsuit$ & 0.775$^\clubsuit$ & 0.819$^\clubsuit$ & 0.808 & 0.846 & 0.819$^\clubsuit$ & 0.874$^\clubsuit$ & 0.762 & 0.825 & 0.782\\

\hdashline

\multirow{2}{*}{\textbf{Sent\MetricName{}-dep} (Ours)}
& $\gamma$ & 0.870 & 0.881 & 0.766 & 0.830 & 0.941$^\clubsuit$ & 0.954$^\clubsuit$ & 0.892 & 0.938 & \underline{0.913}$^\clubsuit$ & \bf{0.918}$^\clubsuit$ & 0.916 & \underline{0.949} & 0.897 \\
& $\rho$   & 0.714 & 0.725 & 0.681 & 0.747 & 0.857$^\clubsuit$ & 0.885$^\clubsuit$ & 0.824 & 0.901 & \bf{0.857}$^\clubsuit$ & \bf{0.912}$^\clubsuit$ & 0.720 & 0.791 & 0.801 \\

\hdashline

\multirow{2}{*}{\textbf{Sent\MetricName{}-ind} (Ours)}
& $\gamma$ & 0.866 & 0.881 & 0.799 & 0.853 & \underline{0.941}$^\clubsuit$ & \underline{0.956}$^\clubsuit$ & \underline{0.915} & 0.952 & \bf{0.915}$^\clubsuit$ & \underline{0.917}$^\clubsuit$ & 0.883 & 0.904 & 0.899 \\
& $\rho$   & 0.709 & 0.720 & 0.681 & 0.747 & \bf{0.879}$^\clubsuit$ & \bf{0.912}$^\clubsuit$ & 0.857 & 0.923 & \underline{0.824}$^\clubsuit$ & \underline{0.885}$^\clubsuit$ & 0.654 & 0.720 & 0.793 \\

\hdashline

\multirow{2}{*}{\textbf{Sent\MetricName{}-sim-dep} (Ours)}
& $\gamma$ & \underline{0.926} & \underline{0.937} & 0.797 & 0.861 & 0.939$^\clubsuit$ & 0.948$^\clubsuit$ & 0.908 & \underline{0.952} & 0.871$^\clubsuit$ & 0.872$^\clubsuit$ & \underline{0.918} & 0.947 & \underline{0.906} \\
& $\rho$   & \bf{0.907} & \bf{0.912} & \underline{0.808} & \underline{0.863} & 0.852$^\clubsuit$ & 0.879$^\clubsuit$ & \bf{0.885} & \underline{0.945} & 0.753$^\clubsuit$ & 0.780$^\clubsuit$ & \bf{0.896} & \bf{0.940} & \bf{0.868} \\

\hdashline

\multirow{2}{*}{\textbf{Sent\MetricName{}-sim-ind} (Ours)}
& $\gamma$ & 0.915 & 0.936 & 0.808 & 0.866 & \bf{0.945}$^\clubsuit$ & \bf{0.956}$^\clubsuit$ & \bf{0.923} & \bf{0.963} & 0.885$^\clubsuit$ & 0.887$^\clubsuit$ & \bf{0.931} & \bf{0.961} & \bf{0.915} \\
& $\rho$   & 0.868 & \underline{0.879} & 0.753 & 0.824 & \underline{0.863}$^\clubsuit$ & \underline{0.901}$^\clubsuit$ & \underline{0.879} & \bf{0.956} & 0.775$^\clubsuit$ & 0.802$^\clubsuit$ & \underline{0.835} & \underline{0.923} & \underline{0.855} \\

\bottomrule
\end{tabular}}
\caption{
Correlation results on GJG15 Ranking. \MetricName{}-sim is based on similarity-based weighting. We highlight the \textbf{highest} scores in bold and the \underline{second-highest} scores with underlines. $\clubsuit$ We exclude unchanged references for higher correlations due to low-quality annotations in some reference sets. Results without excluding references are presented in Appendix~\ref{subsec:exp_full}.}
\label{tab:exp-main-all}
\end{table*}

%% file: tables/SEEDA-base.tex
\begin{table}[tbp!]
\renewcommand{\arraystretch}{1.2}
\renewcommand{\tabcolsep}{4pt}
\resizebox{1.0\linewidth}{!}{
\begin{tabular}{lcccccc}
\toprule

\multirow{2.5}{*}{\textbf{Metric}} 
& \multicolumn{2}{c}{\textbf{SEEDA-S}} 
& \multicolumn{2}{c}{\textbf{SEEDA-E}} 
& \multirow{2.5}{*}{\textbf{Avg.}} \\

\cmidrule(lr){2-3} \cmidrule(lr){4-5}

& \hspace{0.2em}\textbf{$\gamma$}
& \textbf{$\rho$}
& \hspace{0.2em}\textbf{$\gamma$}
& \hspace{0.2em}\textbf{$\rho$} \\

\midrule

\bf{M$^2$} & 0.658 &  0.487 & 0.791 & 0.764 & 0.675 \\ 

\bf{PT-M$^2$} & 0.845 & 0.769 & 0.896 & 0.909 & 0.855 \\ 

\bf{ERRANT} & 0.557 &  0.406 & 0.697 &  0.671 & 0.583 \\ 

\bf{PT-ERRANT} & 0.818 & 0.720 & 0.888 & 0.888 & 0.829 \\ 

\bf{GoToScorer} & \underline{0.929} & 0.881 & 0.901 & 0.937 & 0.912 \\ 

\bf{GLEU} & 0.847 & 0.886 & 0.911 & 0.897 & 0.885 \\ 

\bf{Scribendi Score} & 0.631 & 0.641 & 0.830 & 0.848 & 0.738 \\ 

\bf{SOME} & 0.892 & 0.867 & 0.901 & 0.951 & 0.903 \\ 

\bf{IMPARA} & 0.911 & 0.874 & 0.889 &  0.944 & 0.903 \\ 

\bf{CLEME-dep} & 0.633 & 0.501 & 0.755 & 0.757 & 0.662 \\ 

\bf{CLEME-ind} & 0.616 & 0.466 & 0.736 & 0.708 & 0.632 \\ 

\bf{\MetricName{}-dep (Ours)} & \bf{0.937} & 0.865 & 0.945 & 0.939 & 0.922 \\ 

\bf{\MetricName{}-ind (Ours)} & 0.908 & 0.844 & \bf{0.961} & 0.946 & 0.915 \\ 

\bf{\MetricName{}-sim-dep (Ours)} & 0.923 & \bf{0.914} & 0.948 & \underline{0.974} & \underline{0.940} \\ 

\bf{\MetricName{}-sim-ind (Ours)} & 0.921 & \underline{0.907} & \underline{0.953} & \bf{0.981} & \bf{0.941} \\

\midrule

\bf{Sent-M$^2$} & 0.802 & 0.692 & 0.887 &  0.846 & 0.807 \\ 

\bf{SentERRANT} & 0.758 & 0.643 & 0.860 & 0.825 & 0.772 \\ 

\bf{SentCLEME-dep} & 0.866 & 0.809 & 0.944 & 0.939 & 0.890  \\ 

\bf{SentCLEME-ind} & 0.864 & 0.858 & 0.935 & 0.911 & 0.892  \\  

\bf{Sent\MetricName{}-dep (Ours)} & 0.905 & 0.844 & \underline{0.955} & 0.946 & 0.913 \\ 

\bf{Sent\MetricName{}-ind (Ours)} & 0.875 & 0.837 & 0.953 & 0.953 & 0.905 \\ 

\bf{Sent\MetricName{}-sim-dep (Ours)} & \bf{0.924} & \underline{0.858} & 0.923 & \underline{0.953} & \underline{0.915} \\ 

\bf{Sent\MetricName{}-sim-ind (Ours)} & \underline{0.921} & \bf{0.886} & \bf{0.957} & \bf{0.960} & \bf{0.931} \\ 

\bottomrule
\end{tabular}}
\caption{Results of human correlations on SEEDA Ranking based on TrueSkill (TS).}
\label{tab:seeda-base}
\end{table}

%% file: tables/case-study.tex
\begin{table*}[tbp!]
\renewcommand{\arraystretch}{1.2}
\renewcommand{\tabcolsep}{4pt}
\centering
\resizebox{0.9\linewidth}{!}{
\begin{tabular}{lcccccc}
\toprule

& \bf{Chunk 1}
& \bf{Chunk 2}
& \bf{Chunk 3}
& \bf{Chunk 4}
& \bf{Chunk 5}
& \bf{Chunk 6} \\

\midrule

\bf Source & Do one & who & suffered & from this disease keep it a secret & of infrom & their relatives ? \\

\bf Reference & Does one & who  &  suffers  &  from this disease keep it a secret &  or inform  &  their relatives ?   \\
 
\bf Hypothesis & \FN{\bf Do one} (0.028) & who & \FPne{\bf suffer} (0.011) & from this disease keep it a secret & \FPne{\bf to inform} (0.094) & their relatives ? \\ 

\midrule

\multicolumn{7}{c}{$Hit = 0.00, \quad Wrong = 0.79, \quad Under = 0.21, \quad Over = 0.00$} \\

\bottomrule
\end{tabular}}
\end{table*}

\begin{table*}[tbp!]
\renewcommand{\arraystretch}{1.2}
\renewcommand{\tabcolsep}{3pt}
\resizebox{1.0\linewidth}{!}{
\begin{tabular}{lccccccccc}
\toprule

& \bf{Chunk 1}
& \bf{Chunk 2}
& \bf{Chunk 3}
& \bf{Chunk 4}
& \bf{Chunk 5} 
& \bf{Chunk 6}
& \bf{Chunk 7}
& \bf{Chunk 8}
& \bf{Chunk 9} \\

\midrule





\bf Source & When we are & diagonosed out & with certain genetic & disease & , should we disclose & this result & to & our & relatives ? \\

\bf Ref. & When we are  & diagnosed   & with   certain genetic & diseases & , should we disclose & this result & to & our & relatives ? \\

\bf Hyp. & When we are  &  \FPne{\bf diagnosed out} (0.056)  &  with certain genetic  &  \TP{\bf diseases} (0.006)  &  , should we disclose  &  \FPun{\bf the results} (0.019) &  to  &  \FPun{\bf their} (0.021)  &  relatives ? \\

\midrule

\multicolumn{10}{c}{$Hit = 0.10, \quad Wrong = 0.90, \quad Under = 0.0, \quad Over = 0.39$} \\

\bottomrule
\end{tabular}}
\caption{Study cases of \MetricName{} with similarity-based weighting. We highlight \TP{TP}, \FPne{FP$_{\text{ne}}$}, \FPun{FP$_{\text{un}}$}, and \FN{FN} chunks in different colors. Values in brackets are similarity-based weighting scores.}
\label{tab:case-study}
\end{table*}











%% file: tables/LLM_results.tex
\begin{table}[tbp!]
\renewcommand{\arraystretch}{1.2}
\renewcommand{\tabcolsep}{4pt}
\resizebox{1.0\linewidth}{!}{
\begin{tabular}{lcccccccccc}
\toprule

\multirow{2.5}{*}{\textbf{Dataset}}
& \multicolumn{2}{c}{\textbf{Corpus-EW}}
& \multicolumn{2}{c}{\textbf{Corpus-TS}}
& \multicolumn{2}{c}{\textbf{Sentence-EW}}
& \multicolumn{2}{c}{\textbf{Sentence-TS}} \\ 

\cmidrule(lr){2-3} \cmidrule(lr){4-5} \cmidrule(lr){6-7} \cmidrule(lr){8-9}

& \textbf{$\gamma$}
& \textbf{$\rho$}
& \textbf{$\gamma$}
& \textbf{$\rho$}
& \textbf{$\gamma$}
& \textbf{$\rho$}
& \textbf{$\gamma$}
& \textbf{$\rho$} \\

\midrule

\bf{CoNLL-2014} & 0.697 & 0.659 & 0.759 & 0.720 & 0.626 & 0.654 & 0.696 & 0.698 \\ 

\bf{BN-10GEC} & 0.732 & 0.764 & 0.796 & 0.813 & 0.638 & 0.637 & 0.708 & 0.698 \\ 

\bf{E-Minimal} & 0.709 & 0.786  & 0.779 & 0.819 & 0.642 & 0.692 & 0.715 & 0.747 \\ 

\bf{E-Fluency} & 0.760 & 0.786 & 0.831 & 0.841 & 0.642 & 0.665 & 0.720 & 0.714 \\ 

\bf{NE-Minimal} & 0.777 & 0.823 & 0.839 & 0.861 & 0.654 & 0.747 & 0.723 & 0.791 \\ 

\bf{NE-Fluency} & 0.823 & 0.692 & 0.849 & 0.709  & 0.664 & 0.791 & 0.742 & 0.830 \\

\bottomrule
\end{tabular}}
\caption{Correlation results of LLM-based weighting on GJG15 Ranking.}
\label{tab:llm}
\end{table}

%% file: tables/ablation_disentangled_score.tex
\begin{table}[tbp!]
\renewcommand{\arraystretch}{1.2}
\renewcommand{\tabcolsep}{3pt}
\resizebox{1.0\linewidth}{!}{
\begin{tabular}{lcccccc}
\toprule

\multirow{2.5}{*}{\textbf{Metric}} 
& \multicolumn{2}{c}{\textbf{EW}} 
& \multicolumn{2}{c}{\textbf{TS}} 
& \multirow{2.5}{*}{\textbf{Avg.}} \\

\cmidrule(lr){2-3} \cmidrule(lr){4-5}

& \textbf{$\gamma$}
& \textbf{$\rho$}
& \textbf{$\gamma$}
& \textbf{$\rho$} \\

\midrule

\bf{\MetricName{}-dep-Hit} & \bf 0.599 & 0.593 & \bf 0.673 & \bf 0.648 & \bf 0.628 \\

\bf{\MetricName{}-dep-Wrong} & -0.444 & -0.533 & -0.526 & -0.593 & -0.524 \\

\bf{\MetricName{}-dep-Under} & 0.496 & \bf 0.599 & 0.576 & 0.659 & 0.583 \\

\bf{\MetricName{}-dep-Over} & 0.118 & 0.269 & 0.073 & 0.275  & 0.253 \\





\midrule

\bf{Sent\MetricName{}-dep-Hit} & \bf 0.594 & \bf 0.593 & \bf 0.672 & \bf 0.648 & \bf 0.627 \\

\bf{Sent\MetricName{}-dep-Wrong} & -0.405 & -0.429 & -0.489 & -0.500 & -0.456 \\

\bf{Sent\MetricName{}-dep-Under} & 0.489 & 0.511 & 0.572 & 0.582  & 0.539 \\

\bf{Sent\MetricName{}-dep-Over} & -0.247 & -0.363 & -0.346 & -0.440 & -0.349 \\





\bottomrule
\end{tabular}}
\caption{Correlation results of each disentangled score on GJG15 Ranking.}
\label{tab:disentangled_score}
\end{table}

%% file: tables/efficiency.tex
\begin{table}[tbp!]
\renewcommand{\arraystretch}{1.2}
\renewcommand{\tabcolsep}{4pt}
\centering
\resizebox{0.8\linewidth}{!}{
\begin{tabular}{lc}
\toprule

\textbf{Metric} & \textbf{Time (Seconds)} \\

\midrule

\textbf{ERRANT} & 33.4 \\
\textbf{GLEU} & 21.5 \\
\textbf{CLEME-dep} & 54.1 \\
\textbf{CLEME-ind} & 54.1 \\
\textbf{(Sent)\MetricName{}-dep} & 54.1 \\
\textbf{(Sent)\MetricName{}-ind} & 54.1 \\
\textbf{(Sent)\MetricName{}-sim-ind} & 88.4 \\
\textbf{(Sent)\MetricName{}-sim-ind} & 87.6 \\

\bottomrule
\end{tabular}}
\caption{Efficiency of metrics.}
\label{tab:efficiency}
\end{table}

%% file: chapters/06_conclusion.tex
\section{Conclusion}
This paper introduces \MetricName{}, an interpretable evaluation metric for GEC that effectively highlights four key aspects of systems. By incorporating edit weighting techniques, we overcome the challenges traditional reference-based metrics face in recognizing semantic subtleties. Extensive experiments and analyses confirm the effectiveness and robustness of our method. We anticipate that \MetricName{} will offer a valuable perspective in the GEC community.

\section*{Limitation}
\paragraph{Limitation in languages and datasets.}
While \MetricName{} is adaptable to various languages, its efficiency beyond English remains unverified. Additionally, the reference sets employed in our experiments stem from the CoNLL-2014 shared task, which involves a second language dataset. To confirm the robustness of our methods, it's necessary to conduct further experiments using evaluation datasets that cover a range of languages and text domains. Finally, we highly encourage the creation of new GEC evaluation datasets to foster progress.

\paragraph{Lack of further human evaluation for interpretability.}
The experiments discussed in the paper are primarily concerned with assessing the correlation between automatic metrics and human judgments. However, they fall short of providing a thorough analysis of the method's interpretability. Although we showcase the strong correlation performance of \MetricName{}, its interpretability is still unverified. In future work, we will conduct human evaluation experiments to showcase the interpretability of our method.

\section*{Ethics Statement}
In this paper, we validate the effectiveness and robustness of our proposed approach using the CoNLL-2014, BN-10GEC, and SN-8GEC reference datasets. These datasets are sourced from publicly available resources on legitimate websites and do not contain any sensitive data. Additionally, all the baselines employed in our experiments are publicly accessible GEC metrics, and we have duly cited the respective authors. We confirm that all datasets and baselines utilized in our experiments are consistent with their intended purposes.

\section*{Acknowledgements}
This research is supported by National Natural Science Foundation of China (Grant No. 62276154), Research Center for Computer Network (Shenzhen) Ministry of Education, the Natural Science Foundation of Guangdong Province (Grant No. 2023A1515012914 and 440300241033100801770), Basic Research Fund of Shenzhen City (Grant No. JCYJ20210324120012033, JCYJ20240813112009013 and GJHZ20240218113603006), the Major Key Project of PCL (NO. PCL2024A08).

%% file: chapters/appendix.tex
\appendix

\input{figures/prompt}

\section{LLM-based Edit Weighting}
\label{appendix:llm}

Because of the powerful semantic comprehension abilities of LLMs~\cite{qin2024large,tan2024large,ye2025excgec,yu2024mind,tang2025gmsa,yan2025position,DBLP:conf/nips/LiZLML0HY24,DBLP:conf/sigir/LiLHYS022,DBLP:conf/emnlp/DuW0D0LZVZSZGL024,DBLP:conf/coling/HuangMLHZ0Z24,DBLP:journals/tkde/LiHZZLLCZS23,DBLP:conf/aaai/YuJLHWLCLLTZZXH24,DBLP:conf/iclr/LiLWJZZWZH0Y25,DBLP:journals/corr/abs-2411-17558}, recent studies~\cite{chu2025llm,ye2025position,ye2025corrections,ye2024productagent,hu2024llm,chen2024humans,su2025essayjudge,zou2025revisiting,DBLP:conf/coling/XuLD0CJZLXH25,DBLP:conf/iclr/LiHKLGQTZSY25,DBLP:journals/corr/abs-2403-04247,DBLP:journals/corr/abs-2501-01945,DBLP:conf/aaai/LiL0LHYY024} have generated interest in employing LLMs for text assessment on various NLP tasks. Building on this idea, we use Llama-2-7B~\cite{touvron2023llama} as a scorer to determine edit weights. The prompt for edit weighting is presented in Figure~\ref{fig:llm}. We set the temperature to 0.1 to ensure consistent and certain results. We instruct the LLM to evaluate each edit individually to prevent interference from other grammatical errors. Edit weights vary from 1 to 5, with higher values representing a greater need for correction. We do not specify the types of edits to the LLM; instead, we allow the LLM to directly evaluate the importance of edits through its inherent language understanding abilities. An input is composed of an uncorrected sentence and a certain edit.

\section{Details about GEC Meta-Evaluation}
\label{appendix:GEC_Meta_Evaluation}


\subsection{Human Rankings}
\label{appendix:Human_rankings}

\paragraph{GJG15 ranking.}
\citet{grundkiewicz2015human} propose the first large-scale human judgement dataset for 12 participating systems of the CoNLL-2014 shared task. In this assessment, 8 native speakers are asked to rank the systems' outputs from best to worst. Two system ranking lists are generated using Expected Wins (EW) and TrueSkill (TS), respectively.

\paragraph{SEEDA ranking.}
\citet{kobayashi2024revisiting} identify several limitations of the GJG15 ranking dataset, and propose a new human ranking dataset called SEEDA. SEEDA consists of corrections with human ratings along two different granularities: edit-based and sentence-based, covering 12 state-of-the-art systems, including large language models (LLMs), and two human corrections with different focuses. Three native English speakers participate in the annotation process. Similar to~\citet{grundkiewicz2015human}, the overall human rankings are derived from TrueSkill (TS) and Expected Wins (EW) based on pairwise judgments.

\subsection{Ranking Algorithms}
Our employed human judgments are originally pairwise comparisons, i.e., humans choose the better of two available system outputs. The overall rankings are derived by using ranking algorithms, including Expected Wins (EW) and TrueSkill (TS).

\paragraph{Expected Wins (EW)}
EW~\cite{bojar-etal-2013-findings} is a derived ranking metric that quantifies the theoretical number of wins a participant is expected to achieve against a defined set of opponents. It is calculated by summing the probability of winning against each opponent, where these probabilities are typically derived from an existing skill rating system. EW provides a single aggregate score for ranking, useful for pre-match seeding or assessing theoretical group performance.

\paragraph{TrueSkill (TS)}
TS~\cite{sakaguchi2014efficient} is a Bayesian skill rating system developed by Microsoft Research. Unlike simpler systems, TS models a participant's skill as a probability distribution ($N(\mu, \sigma^2)$), where $\mu$ represents the estimated skill level and $\sigma$ quantifies the uncertainty in that estimate. Upon match outcomes, TS updates these distributions using Bayesian inference, allowing for rapid adjustments and robust ranking. A key advantage is its inherent support for multi-player or team-based matches and the explicit handling of draws. Participants are typically ranked by a conservative estimate of their skill, such as $\mu-3\sigma$, which accounts for confidence.

\input{tables/statistics}

\input{tables/GJG15_sentence_full}

\subsection{Statistics of Reference Datasets}
\label{subsec:statistics}
Table~\ref{tab:statistics} presents the statistics of all the reference sets involved in our experiments.

\subsection{Baseline Metrics}
\label{appendix:Baseline_metrics}
In our evaluation, we compare our method with the following reference-based baseline metrics, including corpus and sentence-level variants:

\begin{itemize}[leftmargin=*]
\item[$\bullet$] \textbf{M$^2$} and \textbf{SentM$^2$}~\cite{dahlmeier-ng-2012-better} dynamically extract the hypothesis edits with the maximum overlap of gold annotations by utilizing the Levenshtein algorithm.

\item[$\bullet$] \textbf{GLEU} and \textbf{SentGLEU}~\cite{napoles2015ground} are BLEU-like GEC metrics based on n-gram matching, rewarding hypothesis n-grams that align with the reference but not the source, while penalizing those aligning solely with the source. GLEU is the main metric in JFLEG, an English GEC dataset that highlights holistic fluency edits.

\item[$\bullet$] \textbf{ERRANT} and \textbf{SentERRANT}~\cite{bryant-etal-2017-automatic} are among the most widely recognized in grammatical error correction. They enhance the accuracy of edit extraction by employing a linguistically refined version of the Damerau-Levenshtein algorithm.

\item[$\bullet$] \textbf{PT-M$^2$} and \textbf{SentPT-M$^2$}~\cite{gong-etal-2022-revisiting} leverage pre-trained language model (PLM) to evaluate GEC systems. The main idea is similar to M$^2$ and ERRANT, but they can leverage the knowledge of pre-trained language models to score edits effectively.

\item[$\bullet$] \textbf{CLEME} and \textbf{SentCLEME}~\cite{ye-etal-2023-cleme} are proposed to provide unbiased scores for multi-reference evaluation. Furthermore, the authors present the correction independence assumption, enabling CLEME to function under either the traditional correction dependence or correction independence assumptions.
\end{itemize}

\input{tables/case4llm}

For the evaluation on SEEDA, we add extra evaluation metrics following the evaluation methods reported in~\citet{kobayashi2024revisiting}:

\begin{itemize}[leftmargin=*]
\item[$\bullet$] \textbf{GoToScorer}~\cite{gotou2020taking}: takes into account the difficulty of error correction when calculating the evaluation score. The difficulty is calculated based on the number of systems that can correct errors.

\item[$\bullet$] \textbf{Scribendi Score}~\cite{islam2021end}: evaluates GEC systems in conjunction with the complexity calculated by GPT-2~\cite{radford2019gpt}, the labeled ranking ratio and the Levenstein distance ratio.

\item[$\bullet$] \textbf{SOME}~\cite{yoshimura2020some}: optimizes human evaluation by fine-tuning BERT separately for criteria such as grammaticality, fluency, and meaning preservation.

\item[$\bullet$] \textbf{IMPARA}~\cite{maeda2022impara}: incorporates a quality assessment model fine-tuned using BERT parallel data and a similarity model that takes into account the effects of editing.
\end{itemize}

\subsection{Details of Determining Trade-off Factors}
\label{appendix:trade_off}
A cross-validation approach was employed on the six reference sets of GJG15 to determine the optimal set. Five of the six reference sets were selected, and an exhaustive exploration of all trade-off factors was conducted. The candidate factors were evaluated at intervals determined by a grid value of 0.05. The optimal factors were then identified and applied to the remaining reference set, yielding resultant corrections. We reiterated this process six times to ascertain the final set of trade-off factors, which exhibited the highest average correction for the remaining reference sets.

\section{Extra Results}
\subsection{Results of Full References}
\label{subsec:exp_full}
The results without excluding unchanged reference sentences are presented in Table~\ref{tab:exp-main-sent-full}. We observe an obvious performance reduction in traditional metrics, especially in NE-Minimal, which contains numerous under-corrections due to annotation by non-experts under the minimal editing guideline. We remove 470 unchanged references in E-Minimal and 612 unchanged references in NE-Minimal. In particular, SentERRANT, SentCLEME-dep, and SentCLEME-ind exhibit negative correlations in NE-Minimal, revealing their lack of robustness. Many metrics also undergo a significant decrease in E-Minimal except \MetricName{}. In the case of E-Minimal, many metrics also show a marked decline, except for \MetricName{}. Our approach achieves the highest or comparable correlations across all reference sets, underscoring its robustness.

\subsection{Case Study of LLM-based Weighting}
\label{app:case_llm}
In Table~\ref{tab:case4llm}, we report instances of \MetricName{} using LLM-based weighting. We notice distinct preferences when comparing similarity-based and LLM-based weighting methods. In the first example, Llama-2 attributes significant weights to Chunks 1 and 2, highlighting key grammatical mistakes. Conversely, it assigns a minor weight to Chunk 5 due to its imperfect modification. The second example shows Llama-2 attributing substantial weights to all chunks. Specifically, for Chunk 2, the hypothesis fails to remove the redundant ``out," emphasizing the under-correction issue. Chunks 6 and 8 display excessive corrections, altering the sentence's original intent and thus indicating considerable over-correction. Generally, Llama-2 tends to ascribe either very high or very low weights to modifications. We speculate it is due to the small scale of the LLM we adopt, impairing its ability to distinguish grammatical errors with varying levels.

\input{tables/CLEME_extra_results}

\subsection{Extra Results of CLEME with Similarity Weighting}
\label{appendix:extra_results_cleme}
We additionally investigate the application of similarity-based weighting to CLEME~\cite{ye-etal-2023-cleme} and present the results on CoNLL2014 in Table~\ref{tab:cleme_extra_results}. We find that similarity-based weighting is superior to length-based weighting for corpus-level CLEME, while the trend is reversed for SentCLEME, and both are better than the unweighted setting.
Moreover, it should be noted that no matter the weighting strategy employed, CLEME consistently underperforms compared to \MetricName{}. This is attributed to the fundamental disparities in design and scoring frameworks between the versions. \MetricName{} was crafted to incorporate these sophisticated weighting techniques, allowing it to better distinguish between diverse error types and deliver a more thorough and refined performance assessment.

\input{tables/average_correlations}

\subsection{Average Correlations.}
\label{appendix:average_correlations}
To analyze our method from a global viewpoint, we present the average correlations derived from the exhaustive enumeration of possible parameter configurations. We explore all potential parameter combinations with increments of 0.05. Table~\ref{tab:average_correlations} shows that all correlations are positive, regardless of the correction assumptions, levels of evaluation, or weighting techniques used. By comparing results from unweighted and similarity-based weighted metrics, we determine that similarity-based weighting substantially enhances human correlation on a global level. Additionally, corpus-level metrics generally achieve higher average values compared to sentence-level metrics. However, sentence-level metrics with optimal parameters can outperform their corpus-level equivalents. This implies that corpus-level metrics might demonstrate greater robustness concerning parameter selection.

\input{tables/CoNLL_results}

\subsection{Details Results on CoNLL-2014}
\label{appendix:detailed_results}
Table~\ref{tab:CoNLL_results} presents a comprehensive evaluation of \MetricName{} on CoNLL-2014 across all GEC systems. Our method offers a clear and quantitative examination of detailed features of GEC systems, which other automatic metrics cannot provide. For instance, the CAMB system attains the top hit-correction score of 0.271 for \MetricName{}-dep, which shows that about 27.1\% of edits by the system are accurate. The wrong-correction score of 0.194 indicates that 19.4\% of edits are correctly placed but incorrect, the under-correction score of 0.534 indicates that 53.4\% of grammatical errors are overlooked by the system, and the over-correction score of 0.470 suggests that 47.0\% of the edits are unnecessary.

As a result, developers and researchers can pinpoint the aspects of their systems that require enhancement. Furthermore, users can select GEC systems that best meet their requirements. For instance, users might opt for a system with a minimal under-correction score in high-stakes situations, as they expect to detect every possible grammatical mistake even though the system might make some unnecessary edits.

%% file: figures/prompt.tex
\begin{figure*}[tb!]
\begin{mdframed}
\setlength{\parindent}{0pt}

\noindent \textit{Prompt}:

As an evaluator for grammatical error correction, you are tasked with assessing the importance of each error. You will be provided with two lines: the first is an uncorrected sentence, the second shows the edit. Then you output the importance score of the given edit.

The scores range from 1 to 5, where a higher score reflects the greater significance of the correction, while a lower score indicates minor importance.

\quad - A score of 1 means the correction is almost negligible and unnecessary.
 
\quad - A score of 2 means the correction has slight influence.

\quad - A score of 3 signifies some impact by the correction.

\quad - A score of 4 means the edit is essential.

\quad - A score of 5 indicates the modification is highly important and necessary. \\

Next, I’ll provide you a sentence with an edit. You should score each edit accordingly. The output should only be the score, with no additional explanation. \\

\textit{Example Input}:

\textit{Uncorrected sentence}: Nowadays the technologies were improved a lot compared to the last century.

\textit{Edit}: were $\to$ have

\textit{Example Output (1-5)}: 5 \\

Note that the output must be a number between 1 and 5. Here is the formal input:

\textit{Uncorrected sentence}: \{uncorrected sentence\}

\textit{Edit}: \{edit\}

\textit{Example Output (1-5)}:

\end{mdframed}
\caption{Prompt of LLM-based weighting.} 
\label{fig:llm}
\end{figure*}

%% file: tables/statistics.tex
\begin{table*}[tb!]
\renewcommand{\arraystretch}{1.2}
\renewcommand{\tabcolsep}{4pt}
\resizebox{1.0\linewidth}{!}{
\begin{tabular}{lcccccc}
\toprule

\bf{Item}
& \bf{CoNLL-2014}
& \bf{BN-10GEC}
& \bf{E-Minimal}
& \bf{E-Fluency}
& \bf{NE-Minimal}
& \bf{NE-Fluency} \\

\midrule

\# \bf Sentence (Length) & 1,312 (23.0) & 1,312 (23.0) & 1,312 (23.0) & 1,312 (23.0) & 1,312 (23.0) & 1,312 (23.0) \\
\# \bf Reference (Length) & 2,624 (22.8) & 13,120 (22.9) & 2,624 (23.2) & 2,624 (22.8) & 2,624 (23.0) & 2,624 (22.2) \\
\# \bf Edit (Length) & 5,937 (1.0) & 36,677 (1.0) & 4,500 (1.0) & 8,373 (1.1) & 4,964 (0.9) & 11,033 (1.2) \\
\# \bf Unchanged Chunk (Length) & 11,174 (4.8) & 93,496 (2.5) & 8,887 (6.3) & 12,823 (3.8) & 10,748 (5.1) & 14,086 (2.9) \\
\# \bf Corrected/Dummy Chunk (Length) & 4,994 (1.3) & 26,948 (2.4) & 3,963 (1.2) & 6,305 (1.7) & 4,221 (1.2) & 6,892 (2.6) \\

\bottomrule
\end{tabular}}
\caption{Statistics of CoNLL-2014~\cite{ng2014conll}, BN-10GEC~\cite{bryant-ng-2015-far} and SN-8GEC~\cite{sakaguchi-etal-2016-reassessing} reference sets. We leverage ERRANT~\cite{bryant-etal-2017-automatic} for edit extraction, and CLEME~\cite{ye-etal-2023-cleme} for chunk extraction.}
\label{tab:statistics}
\end{table*}

%% file: tables/GJG15_sentence_full.tex

\begin{table*}[tb!]
\renewcommand{\arraystretch}{1.2}
\renewcommand{\tabcolsep}{4pt}
\resizebox{1.0\linewidth}{!}{
\begin{tabular}{lclllllllllllll}
\toprule

\multirow{2.5}{*}{\textbf{Metric}} & &
\multicolumn{2}{c}{\textbf{CoNLL-2014}} &
\multicolumn{2}{c}{\textbf{BN-10GEC}} &
\multicolumn{2}{c}{\textbf{E-Minimal}} & 
\multicolumn{2}{c}{\textbf{E-Fluency}} &
\multicolumn{2}{c}{\textbf{NE-Minimal}} &
\multicolumn{2}{c}{\textbf{NE-Fluency}} &
\multicolumn{1}{c}{\multirow{2}{*}{\textbf{Avg.}}} \\

\cmidrule(lr){3-4} \cmidrule(lr){5-6} \cmidrule(lr){7-8} \cmidrule(lr){9-10} \cmidrule(lr){11-12} \cmidrule(lr){13-14} 

& 
& \multicolumn{1}{c}{\textbf{EW}} & \multicolumn{1}{c}{\textbf{TS}}
& \multicolumn{1}{c}{\textbf{EW}} & \multicolumn{1}{c}{\textbf{TS}}
& \multicolumn{1}{c}{\textbf{EW}} & \multicolumn{1}{c}{\textbf{TS}}
& \multicolumn{1}{c}{\textbf{EW}} & \multicolumn{1}{c}{\textbf{TS}}
& \multicolumn{1}{c}{\textbf{EW}} & \multicolumn{1}{c}{\textbf{TS}}
& \multicolumn{1}{c}{\textbf{EW}} & \multicolumn{1}{c}{\textbf{TS}}
\\

\midrule



\multirow{2}{*}{\textbf{SentGLEU}}
& $\gamma$ & 0.784 & 0.828 & 0.756 & 0.826 & 0.624 & 0.581 & 0.785 & 0.846 & 0.218 & 0.142 & 0.778 & 0.848 & 0.668 (\Decrease{0.120}) \\
& $\rho$ & 0.720 & 0.775 & 0.769 & 0.824 & 0.599 & 0.593 & 0.791 & 0.846 & 0.220 & 0.170 & 0.768 & 0.846 & 0.660 (\Decrease{0.131}) \\

\hdashline

\multirow{2}{*}{\textbf{SentERRANT} }
& $\gamma$ & 0.870 & 0.846 & \underline{0.885} & \underline{0.896} & 0.760 & 0.692 & 0.806 & 0.732 & 0.104 & -0.066 & 0.793 & 0.847 & 0.680 (\Decrease{0.130}) \\
& $\rho$ & 0.742 & 0.747 & 0.786 & 0.830 & 0.626 & 0.588 & 0.813 & 0.764 & -0.003 & -0.137 & 0.830 & 0.857 & 0.620 (\Decrease{0.179}) \\

\hdashline





\multirow{2}{*}{\textbf{SentCLEME-dep}}
& $\gamma$ & 0.876 & 0.844 & \bf{0.915} & \bf{0.913} & 0.602 & 0.507 & 0.849 & 0.886 & -0.021 & -0.127 & 0.876 & 0.921 & 0.670 (\Decrease{0.185}) \\
& $\rho$ & 0.824 & 0.808 & \bf{0.835} & \bf{0.874} & 0.451 & 0.412 & 0.824 & 0.863 & -0.181 & -0.247 & 0.791 & 0.846 & 0.592 (\Decrease{0.233}) \\

\hdashline

\multirow{2}{*}{\textbf{SentCLEME-ind} }
& $\gamma$ & 0.868 & 0.857 & 0.539 & 0.453 & 0.513 & 0.410 & 0.841 & 0.877 & -0.061 & -0.181 & 0.852 & 0.896 & 0.572 (\Decrease{0.279}) \\
& $\rho$ & 0.725 & 0.758 & 0.209 & 0.143 & 0.368 & 0.335 & 0.808 & 0.846 & -0.167 & -0.247 & 0.762 & 0.825 & 0.447 (\Decrease{0.335}) \\

\hdashline

\multirow{2}{*}{\textbf{Sent\MetricName{}-dep} (Ours)}
& $\gamma$ & 0.870 & 0.881 & 0.766 & 0.830 & \underline{0.937} & \underline{0.928} & 0.892 & 0.938 & 0.634 & 0.571 & 0.916 & \underline{0.949} & 0.843 (\Decrease{0.054}) \\
& $\rho$ & 0.714 & 0.725 & 0.681 & 0.747 & \bf 0.846 & \bf 0.852 & 0.824 & 0.901 & 0.368 & 0.352 & 0.720 & 0.791 & 0.710 (\Decrease{0.091}) \\

\hdashline

\multirow{2}{*}{\textbf{Sent\MetricName{}-ind} (Ours)}
& $\gamma$ & 0.866 & 0.881 & 0.799 & 0.853 & \bf 0.940 & \bf 0.933 & \underline{0.915} & 0.952 & \underline{0.693} & \underline{0.631} & 0.883 & 0.904 & 0.854 (\Decrease{0.045}) \\
& $\rho$ & 0.709 & 0.720 & 0.681 & 0.747 & \underline{0.819} & \underline{0.835} & 0.857 & 0.923 & 0.423 & 0.401 & 0.654 & 0.720 & 0.707 (\Decrease{0.086}) \\

\hdashline

\multirow{2}{*}{\textbf{Sent\MetricName{}-sim-dep} (Ours)}
& $\gamma$ & \underline{0.926} & \underline{0.937} & 0.797 & 0.861 & 0.914 & 0.902 & 0.908 & \underline{0.952} & 0.607 & 0.550 & \underline{0.918} & 0.947 & \underline{0.852} (\Decrease{0.054}) \\
& $\rho$ & \bf{0.907} & \bf{0.912} & \underline{0.808} & \underline{0.863} & 0.808 & 0.813 & \bf{0.885} & \underline{0.945} & \underline{0.527} & \underline{0.505} & \bf{0.896} & \bf{0.940} & \bf{0.817} (\Decrease{0.051}) \\

\hdashline

\multirow{2}{*}{\textbf{Sent\MetricName{}-sim-ind} (Ours)}
& $\gamma$ & 0.915 & 0.936 & 0.808 & 0.866 & 0.922 & 0.916 & \bf{0.923} & \bf{0.963} & \bf 0.720 & \bf 0.669 & \bf{0.931} & \bf{0.961} & \bf{0.877} (\Decrease{0.038}) \\

& $\rho$ & 0.868 & \underline{0.879} & 0.753 & 0.824 & 0.808 & 0.841 & \underline{0.879} & \bf{0.956} & \bf 0.544 & \bf 0.527 & \underline{0.835} & \underline{0.923} & \underline{0.803} (\Decrease{0.052}) \\

\bottomrule
\end{tabular}}
\caption{Correlation results on GJG15 Ranking. We report the results without excluding unchanged reference sentences and the reduction compared with Table~\ref{tab:exp-main-all}. We highlight the \textbf{highest} scores in bold and the \underline{second-highest} scores with underlines.}
\label{tab:exp-main-sent-full}
\end{table*}

%% file: tables/case4llm.tex
\begin{table*}[tbp!]
\renewcommand{\arraystretch}{1.2}
\renewcommand{\tabcolsep}{4pt}
\centering
\resizebox{0.9\linewidth}{!}{
\begin{tabular}{lcccccc}
\toprule

& \bf{Chunk 1}
& \bf{Chunk 2}
& \bf{Chunk 3}
& \bf{Chunk 4}
& \bf{Chunk 5}
& \bf{Chunk 6} \\

\midrule

\bf Source & Do one & who & suffered & from this disease keep it a secret & of infrom & their relatives ? \\

\bf Reference & Does one & who & suffers & from this disease keep it a secret & or inform & their relatives ?   \\
 
\bf Hypothesis & \FN{\bf Do one} (5) & who & \FPne{\bf suffer} (5) & from this disease keep it a secret & \FPne{\bf to inform} (1) & their relatives ?  \\

\midrule

\multicolumn{7}{c}{$Hit = 0.00, \quad Wrong = 0.55, \quad Under = 0.45, \quad Over = 0.00$} \\

\bottomrule
\end{tabular}}
\end{table*}

\begin{table*}[tbp!]
\renewcommand{\arraystretch}{1.2}
\renewcommand{\tabcolsep}{4pt}
\resizebox{1.0\linewidth}{!}{
\begin{tabular}{lccccccccc}
\toprule

& \bf{Chunk 1}
& \bf{Chunk 2}
& \bf{Chunk 3}
& \bf{Chunk 4}
& \bf{Chunk 5} 
& \bf{Chunk 6}
& \bf{Chunk 7}
& \bf{Chunk 8}
& \bf{Chunk 9} \\

\midrule





\bf Source & When we are & diagonosed out & with certain genetic & disease & , should we disclose & this result & to & our & relatives ? \\

\bf Ref. & When we are  & diagnosed   & with   certain genetic & diseases & , should we disclose & this result & to & our & relatives ? \\

\bf Hyp. & When we are & \FPne{\bf diagnosed out} (5) & with certain genetic & \TP{\bf diseases} (5) & , should we disclose & \FPun{\bf the results} (4) & to & \FPun{\bf their} (5) & relatives ? \\

\midrule

\multicolumn{10}{c}{$Hit = 0.50, \quad Wrong = 0.50, \quad Under = 0.00, \quad Over = 0.47 $} \\

\bottomrule
\end{tabular}}
\caption{Study cases of \MetricName{} with LLM-based weighting. We highlight \TP{TP}, \FPne{FP$_{\text{ne}}$}, \FPun{FP$_{\text{un}}$}, and \FN{FN} chunks in different colors. Values in brackets are LLM-based weighting scores.}
\label{tab:case4llm}
\end{table*}











%% file: tables/CLEME_extra_results.tex
\begin{table}[tb!]
\renewcommand{\arraystretch}{1.1}
\renewcommand{\tabcolsep}{4pt}
\resizebox{1.0\linewidth}{!}{
\begin{tabular}{lcccccc}
\toprule

\multirow{2.5}{*}{\textbf{Metric}}
& \multicolumn{2}{c}{\textbf{EW}}
& \multicolumn{2}{c}{\textbf{TS}} \\

\cmidrule(lr){2-3} \cmidrule(lr){4-5}

& \textbf{$\gamma$}
& \textbf{$\rho$}
& \textbf{$\gamma$}
& \textbf{$\rho$} \\

\midrule

\bf{CLEME-dep-unw} & 0.638 & 0.654 & 0.681 & 0.709 \\ 
\bf{CLEME-ind-unw} & 0.640 & 0.648 & 0.680 & 0.698 \\ 
\bf{CLEME-dep-len} & \bf 0.700 & 0.665 & 0.691 & 0.742 \\ 
\bf{CLEME-ind-len} & 0.649 & 0.709 & 0.691 & 0.731 \\ 
\bf{CLEME-dep-sim} & 0.655 & \bf 0.764 & \bf 0.698 & \bf 0.797 \\ 
\bf{CLEME-ind-sim} & 0.641 & 0.720 & 0.687 & 0.747 \\

\midrule

\bf{SentCLEME-dep-unw} & 0.853 & 0.687 & 0.805 & 0.604 \\
\bf{SentCLEME-ind-unw} & 0.790 & 0.275 & 0.722 & 0.181 \\
\bf{SentCLEME-dep-len} & 0.876 & \bf 0.824 & 0.844 & \bf 0.808 \\
\bf{SentCLEME-ind-len} & 0.868 & 0.725 & \bf 0.857 & 0.758 \\
\bf{SentCLEME-dep-sim} & \bf 0.888 & 0.692 & 0.844 & 0.648 \\
\bf{SentCLEME-ind-sim} & 0.843 & 0.500 & 0.786 & 0.434 \\

\bottomrule
\end{tabular}}
\caption{Extra results of CLEME with different edit weighting techniques: unweighting (unw), length-based weighting (len), and similarity-based weighting (sim).}
\label{tab:cleme_extra_results}
\end{table}

%% file: tables/average_correlations.tex
\begin{table}[tbp!]
\renewcommand{\arraystretch}{1.2}
\renewcommand{\tabcolsep}{4pt}
\resizebox{1.0\linewidth}{!}{
\begin{tabular}{lcccccc}
\toprule

\multirow{2.5}{*}{\textbf{Metric}}
& \multicolumn{2}{c}{\textbf{EW}}
& \multicolumn{2}{c}{\textbf{TS}}
& \multirow{2.5}{*}{\textbf{Avg.}} \\

\cmidrule(lr){2-3} \cmidrule(lr){4-5}

& \textbf{$\gamma$}
& \textbf{$\rho$}
& \textbf{$\gamma$}
& \textbf{$\rho$} \\

\midrule

\bf{\MetricName{}-dep} & 0.461 & 0.423 & 0.483 & 0.457 & 0.456 \\ 

\bf{\MetricName{}-ind} & 0.468 & 0.421 & 0.489 & 0.453 & 0.458 \\ 

\bf{\MetricName{}-sim-dep} & 0.559 & 0.592 & 0.581 & \bf{0.624} & 0.589 \\ 

\bf{\MetricName{}-sim-ind} & \bf{0.566} & \bf{0.593} & \bf{0.588} & 0.622  & \bf{0.592} \\ 

\midrule

\bf{Sent\MetricName{}-dep} & 0.374 &  0.305 & 0.362 & 0.290 & 0.333 \\ 

\bf{Sent\MetricName{}-ind} & 0.372 & 0.302 & 0.356 & 0.283 & 0.328 \\ 

\bf{Sent\MetricName{}-sim-dep} & 0.410 & \bf{0.361} & \bf{0.400} & \bf{0.345}  & \bf{0.379} \\ 

\bf{Sent\MetricName{}-sim-ind} & \bf{0.412} & 0.360 & 0.399 & 0.338 & 0.377 \\ 

\bottomrule
\end{tabular}}
\caption{Average correlations of (Sent)\MetricName{} and (Sent)\MetricName{}-sim on CoNLL-2014.}
\label{tab:average_correlations}
\end{table}

%% file: tables/CoNLL_results.tex
\begin{table*}[tbp!]
\renewcommand{\arraystretch}{1.0}
\renewcommand{\tabcolsep}{4pt}
\centering
\scalebox{0.62}{
\begin{tabular}{llccccccccccccc}

\toprule
\bf{Metric} &
& \multicolumn{1}{c}{\bf{AMU}}
& \multicolumn{1}{c}{\bf{CAMB}}
& \multicolumn{1}{c}{\bf{CUUI}}
& \multicolumn{1}{c}{\bf{IITB}}
& \multicolumn{1}{c}{\bf{INPUT}}
& \multicolumn{1}{c}{\bf{IPN}}
& \multicolumn{1}{c}{\bf{NTHU}}
& \multicolumn{1}{c}{\bf{PKU}}
& \multicolumn{1}{c}{\bf{POST}}
& \multicolumn{1}{c}{\bf{RAC}}
& \multicolumn{1}{c}{\bf{SJTU}}
& \multicolumn{1}{c}{\bf{UFC}}
& \multicolumn{1}{c}{\bf{UMC}} \\

\midrule

\multirow{22}{*}{\bf{\MetricName{}-dep}}
& \bf{TP} & 380 & 584 & 471 & 22 & 0 & 39 &  330 & 246 & 412 & 254 & 85 & 32 & 260  \\
& \quad sim & 9.20 & 12.66 & 7.58 & 0.39 & 0.00 & 0.77 & 5.79 & 6.69 & 8.80 & 6.68 & 1.50 & 0.42 & 5.29 \\
& \bf{FP} & 817 & 1307 & 964 & 67 & 0 & 488 & 905 & 709 & 1145 & 782 & 272 & 18 & 789 \\
& \quad sim & 16.03 & 30.92 & 16.06 & 1.80 & 0.00 & 11.93 & 24.56 & 14.36 & 19.25 & 11.98 & 6.49 & 0.25 & 18.26 \\
& \bf{FP$_{ne}$} & 276 & 418 & 311 & 34 & 0 & 149 & 302 & 254 & 316 & 259 & 76 & 12 & 245 \\
& \quad sim & 4.08 & 6.55 & 3.68 & 0.75 & 0.00 & 4.61 & 5.89 & 4.06 & 4.60 & 3.80 & 2.30 & 0.17 & 3.83 \\
& \bf{FP$_{un}$} & 541 & 889 & 653 & 33 & 0 & 339 & 603 & 455 & 829 & 523 & 196 & 6 & 544 \\
& \quad sim & 11.95 & 24.36 & 12.38 & 1.05 & 0.00 & 7.33 & 18.67 & 10.30 & 14.64 & 8.18 & 4.19 & 0.08 & 14.43 \\
& \bf{FN} & 1360 & 1150 & 1357 & 2057 & 1782 & 2886 & 1388 & 1454 & 1354 & 1487 & 1668 & 2087 & 1461 \\
& \quad sim & 34.25 & 28.45 & 36.21 & 78.39 & 48.24 & 83.10 & 46.53 & 36.15 & 34.48 & 38.00 & 56.28 & 51.27 & 39.60 \\


& \bf{Hit} & 0.188 & 0.271 & 0.220 & 0.010 & 0.00 & 0.013 & 0.163 & 0.126 & 0.198 & 0.127 & 0.046 & 0.015 &  0.132 \\
& \quad sim & 0.194 & 0.266 & 0.160 & 0.005 & 0.00 & 0.009 & 0.100 & 0.143 & 0.184 & 0.138 & 0.025 & 0.008 & 0.109 \\
& \bf{Wrong} & 0.137 & 0.194 & 0.145 & 0.016 & 0.00 & 0.048 & 0.150 & 0.130 & 0.152 & 0.130 & 0.042 & 0.006 & 0.125 \\
& \quad sim & 0.086 & 0.138 & 0.078 & 0.009 & 0.00 & 0.052 & 0.101 & 0.0866 & 0.096 & 0.078 & 0.038 &  0.003 &  0.079 \\
& \bf{Under} & 0.675 & 0.534 & 0.634 & 0.973 & 1.00 & 0.939 & 0.687 & 0.744 & 0.650 & 0.744 & 0.912 & 0.979 & 0.743 \\
& \quad sim & 0.721 & 0.597 & 0.763 & 0.986 & 1.00 & 0.939 & 0.799 & 0.771 & 0.720 & 0.784 & 0.937 &  0.989 &  0.813 \\
& \bf{Over} & 0.452 & 0.470 & 0.455 & 0.371 & 0.00 & 0.643 & 0.488 &  0.476 & 0.532 & 0.505 & 0.549 & 0.12 & 0.519 \\
& \quad sim & 0.474 & 0.559 & 0.524 & 0.478 & 0.00 & 0.577 & 0.615 & 0.490 & 0.522 & 0.438 & 0.524 & 0.116 &  0.613 \\
& \bf{Score} & 0.483 & 0.508 & 0.497 & 0.431 & 0.45 & 0.408 & 0.463 & 0.450 & 0.479 & 0.505 & 0.434 &  0.450 & 0.453 \\
& \quad sim & 0.503 & 0.520 & 0.484 & 0.425 & 0.45 & 0.408 & 0.439 & 0.474 & 0.491 & 0.438 &  0.424 & 0.448 & 0.452 \\

\midrule

\multirow{22}{*}{\bf{Sent\MetricName{}-dep}}
& \bf{TP} & 376 & 580 & 467 & 22 & 0 & 39 &  327 & 244 & 409 & 251 & 84 & 32 & 259 \\
& \quad sim & 9.14 & 12.63 & 7.52 & 0.39 & 0.00 & 0.76 &  5.72 & 6.65 & 8.75 & 6.59 & 1.48 & 0.42 & 5.23 \\
& \bf{FP} & 821 & 1311 & 968 & 67 & 0 & 488 & 
 908 & 711 & 1148 & 785 & 273 & 18 & 790 \\
& \quad sim & 16.49 & 31.25 & 16.50 & 1.85 & 0.00 & 13.00 & 24.83 & 14.38 & 19.36 & 12.34 &  7.13 & 0.26 &  18.47 \\
& \bf{FP$_{ne}$} & 286 & 431 & 320 & 22 & 0 & 132 & 310 & 262 & 326 & 271 & 81 & 10 & 255 \\
& \quad sim & 4.60 & 7.51 & 4.27 & 0.44 & 0.00 & 2.62 & 6.58 & 4.58 & 5.06 & 4.02 & 1.28 & 0.15 & 4.39 \\
& \bf{FP$_{un}$} & 535 & 880 & 648 & 45 & 0 & 356 & 598 & 449 & 822 & 514 & 192 & 8 & 535 \\
& \quad sim & 11.89 & 23.74 & 12.23 & 1.42 & 0.00 & 10.39 & 18.24 &  9.80 & 14.30 & 8.32 & 5.85 & 0.12 & 14.07 \\
& \bf{FN} & 1600 & 1374 & 1577 & 1972 & 1982 & 1940 & 1660 & 1712 & 1587 & 1744 & 1900 & 1980 & 1714 \\
& \quad sim & 43.65 & 35.92 & 45.22 & 57.46 & 58.31 & 54.69 & 46.92 & 46.02 & 43.09 & 46.05 & 55.32 & 58.35 & 48.02 \\


& \bf{Hit} & 0.136 & 0.210 & 0.163 & 0.008 & 0.00 & 0.013 & 0.119 & 0.088 & 0.142 &  0.089 & 0.032 & 0.012 & 0.091 \\
& \quad sim & 0.131 & 0.205 & 0.142 & 0.007 & 0.00 & 0.011 &  0.104 & 0.088 & 0.129 & 0.086 & 0.027 & 0.008 & 0.087 \\
& \bf{Wrong} & 0.080 & 0.129 & 0.090 & 0.005 & 0.00 & 0.038 & 0.095 & 0.076 & 0.088 &  0.071 & 0.023 & 0.002 & 0.070 \\
& \quad sim & 0.063 & 0.102 & 0.066 & 0.004 & 0.00 & 0.033 & 0.079 & 0.059 & 0.070 & 0.051 & 0.020 &  0.001 & 0.059 \\
& \bf{Under} & 0.500 & 0.392 & 0.479 &  0.675 & 0.687 & 0.639 & 0.496 & 0.538 & 0.486 & 0.551 & 0.637 & 0.678 & 0.546 \\
& \quad sim & 0.519 & 0.419 & 0.517 & 0.673 & 0.684 & 0.645 & 0.524 & 0.553 & 0.509 &  0.567 &  0.641 & 0.680 & 0.557 \\
& \bf{Over} & 0.248 & 0.419 & 0.293 & 0.031 & 0.00 & 0.242 & 0.304 & 0.235 & 0.342 & 0.232 & 0.121 & 0.006 & 0.267 \\
& \quad sim & 0.241 & 0.421 & 0.294 & 0.030 & 0.00 & 0.224 & 0.302 & 0.224 & 0.331 & 0.203 & 0.119 & 0.005 & 0.267 \\
& \bf{Score} & 0.498 & 0.513 & 0.507 & 0.467 & 0.466 & 0.447 & 0.481 & 0.475 & 0.495 & 0.477 & 0.469 & 0.471 & 0.476 \\
& \quad sim & 0.502 & 0.520 & 0.504 & 0.467 & 0.466 & 0.449 & 0.479 & 0.481 & 0.494 & 0.484 & 0.467 & 0.469 & 0.479 \\

\midrule
\midrule

\multirow{21}{*}{\bf{\MetricName{}-ind}}
& \bf{TP} & 388 & 596 & 487 & 22 & 0 & 39 & 338 & 248 & 420 & 255 & 85 & 32 & 262 \\
& \quad sim & 9.47 & 13.11 & 7.99 & 0.40 & 0.00 & 0.81 & 6.13 & 6.80 & 9.07 & 6.91 & 1.54 & 0.47 & 5.49 \\
& \bf{FP} & 809 & 1295 & 948 & 67 & 0 & 488 &  897 & 707 & 1137 & 781 & 272 & 18 & 787 \\
& \quad sim & 14.74 & 28.11 & 14.42 & 1.91 & 0.00 & 11.82 & 22.93 & 13.03 & 17.62 & 11.23 &  6.46 & 0.25 & 16.99 \\

& \bf{FP$_{ne}$} & 408 & 627 & 449 & 34 & 0 & 234 & 447 &  388 & 487 & 406 & 134 & 12 & 366 \\
& \quad sim & 6.32 & 10.62 & 5.51 & 0.86 & 0.00 & 4.79 & 9.50 & 7.30 &  7.12 & 5.56 &  2.41 & 0.17 & 6.14 \\
& \bf{FP$_{un}$} & 401 & 668 & 499 & 33 & 0 & 254 & 450 & 319 & 650 & 375 & 138 & 6 & 421 \\
& \quad sim & 8.42 & 17.49 &  8.91 & 1.05 & 0.00 & 7.03 & 13.43 &  5.73 & 10.50 & 5.67 & 4.05 & 0.08 &  10.85 \\

& \bf{FN} & 1029 & 778 & 984 & 1497 & 1530 & 1382 & 1045 & 1129 & 989 & 1135 & 1398 & 1506 & 1136 \\
& \quad sim & 26.88 & 20.31 & 27.94 & 53.23 & 41.31 & 50.21 & 36.83 & 28.40 & 26.59 &  29.30 & 40.63 & 41.49 & 31.88 \\



& \bf{Hit} & 0.213 & 0.298 & 0.254 & 0.014 & 0.000 & 0.024 & 0.185 & 0.141 & 0.222 & 0.142 & 0.053 & 0.021 & 0.149 \\
& \quad sim &  0.222 & 0.298 &  0.193 & 0.007 & 0.000 & 0.015 & 0.117 & 0.160 & 0.212 & 0.165 & 0.035 & 0.011 & 0.126 \\
& \bf{Wrong} & 0.224 &  0.313 & 0.234 & 0.022 & 0.000 & 0.141 & 0.244 &  0.220 & 0.257 & 0.226 & 0.083 & 0.008 & 0.207 \\
& \quad sim &  0.148 & 0.241 & 0.133 & 0.016 & 0.000 & 0.086 & 0.181 & 0.172 & 0.166 & 0.133 & 0.054 &  0.004 &  0.141 \\
& \bf{Under} &  0.564 & 0.389 & 0.513 & 0.964 & 1.000 & 0.835 & 0.571 & 0.640 & 0.522 & 0.632 & 0.865 & 0.972 &  0.644 \\
& \quad sim & 0.630 & 0.461 & 0.674 & 0.977 & 1.000 & 0.900 &  0.702 & 0.668 & 0.622 & 0.701 & 0.911 & 0.985 & 0.733 \\
& \bf{Over} & 0.335 & 0.353 & 0.348 & 0.371 & 0.000 & 0.482 & 0.364 & 0.334 & 0.417 & 0.362 & 0.387 & 0.12 & 0.401 \\
& \quad sim &  0.348 &  0.424 & 0.397 & 0.454 & 0.000 & 0.557 &  0.462 & 0.289 & 0.393 & 0.313 & 0.506 & 0.11 & 0.483 \\
& \bf{Score} & 0.472 & 0.486 & 0.490 & 0.432 & 0.450 & 0.389 & 0.448 & 0.434 & 0.461 & 0.431 & 0.431 &  0.453 & 0.439 \\
& \quad sim & 0.503 &  0.508 &  0.490 & 0.426 & 0.450 &  0.400 & 0.428 & 0.463 & 0.489 & 0.479 &  0.425 &  0.449 & 0.446 \\

\midrule

\multirow{15}{*}{\bf{Sent\MetricName{}-ind}}
& \bf{TP}-sim & 9.16 & 12.59 & 7.73 & 0.40 & 0.00 & 0.75 &  5.93 & 6.67 & 8.77 & 6.67 & 1.50 & 0.47 & 5.21 \\
& \bf{FP}-sim & 15.83 & 29.93 & 15.62 & 1.76 & 0.00 & 12.58 & 24.30 & 14.17 & 18.94 & 12.00 & 6.84 &  0.27 & 17.76 \\

& \bf{FP$_{ne}$}-sim &   7.20 & 12.38 & 6.58 & 0.70 & 0.00 & 5.27 & 10.94 &  8.38 & 8.37 & 6.25 & 2.70 & 0.19 & 6.81 \\
& \bf{FP$_{un}$}-sim & 8.63 & 17.54 & 9.03 & 1.07 & 0.00 & 7.31 & 13.36 &  5.80 & 10.57 & 5.75 & 4.14 & 0.08 & 10.95 \\

& \bf{FN}-sim & 31.54 & 22.55 & 32.06 & 47.73 & 48.90 & 43.66 &33.43 & 33.87 &  30.37 & 33.61 & 45.12 & 48.29 & 36.24 \\


& \bf{Hit} & 0.155 & 0.239 &  0.189 & 0.010 & 
 0.000 & 0.016 & 0.137 & 0.100 & 0.165 &  0.106 & 0.036 & 0.015 &  0.105 \\
& \quad sim & 0.154 & 0.240 & 0.174 & 0.009 &  0.000 &  0.014 & 0.125 & 0.100 &  0.155 & 0.103 & 0.033 &  0.012 & 0.102 \\
& \bf{Wrong} & 0.159 & 0.261 & 0.178 & 0.015 &  0.000 &  0.110 & 0.192 & 0.165 & 0.192 & 0.162 &  0.059 & 0.005 & 0.147 \\
& \quad sim & 0.134 &  0.229 & 0.147 & 0.013 &  0.000 & 0.094 & 0.170 & 0.144 & 0.164 & 0.129 & 0.051 & 0.004 &  0.127 \\
& \bf{Under} & 0.403 & 0.268 & 0.373 & 0.627 & 0.647 & 0.563 & 0.390 & 0.447 & 0.375 & 0.450 & 0.574 & 0.635 & 0.449 \\
& \quad sim & 0.429 & 0.299 & 0.415 & 0.629 & 
 0.647 & 0.580 & 0.425 & 0.467 & 0.407 & 0.475 & 0.586 & 0.639 & 0.471 \\
& \bf{Over} & 0.183 &  0.315 & 0.227 & 0.023 &  0.000 & 0.171 & 0.224 & 0.163 & 0.266 & 0.165 & 0.086 &  0.004 &  0.206 \\
& \quad sim &  0.183 & 0.320 & 0.230 & 0.023 &  0.000 & 0.169 & 0.229 & 0.159 & 0.264 & 0.150 & 0.089 & 0.005 & 0.211 \\
& \bf{Score} & 0.485 & 0.486 & 0.493 & 0.466 & 0.468 & 0.428 & 0.461 & 0.453 & 0.474 & 0.458 &  0.461 &  0.474 & 0.461 \\
& \quad sim & 0.493 & 0.498 & 0.496 & 0.466 & 0.468 & 0.432 & 0.462 & 0.461 &  0.478 & 0.469 &  0.462 & 0.473 & 0.466 \\

\bottomrule
\end{tabular}}
\caption{\label{tab:CoNLL_results}
Detailed evaluation results across GEC systems on CoNLL-2014. We report True Positives (TPs), False Positives (FPs), False Negatives (FNs), and True Negatives (TNs) with or w/o similarity-based weighting (sim).}
\end{table*}